\documentclass{article}
\usepackage{ijcai21}

\usepackage{times}
\usepackage{soul}
\usepackage{url}
\usepackage[hidelinks]{hyperref}
\usepackage[utf8]{inputenc}
\usepackage[small]{caption}
\usepackage{graphicx}
\usepackage{amsmath}
\usepackage{amsthm}
\usepackage{booktabs}
\usepackage{algorithm}
\usepackage{algorithmic}
\usepackage{array}
\usepackage{threeparttable}
\usepackage{xpatch}
\usepackage{titlesec}

\usepackage{multirow}
\usepackage{multicol}
\usepackage{amssymb}
\newcommand{\tabincell}[2]{\begin{tabular}{@{}#1@{}}#2\end{tabular}}

\makeatletter
\DeclareRobustCommand{\cev}[1]{%
  \mathpalette\do@cev{#1}%
}
\newcommand{\do@cev}[2]{%
  \fix@cev{#1}{+}%
  \reflectbox{$\m@th#1\vec{\reflectbox{$\fix@cev{#1}{-}\m@th#1#2\fix@cev{#1}{+}$}}$}%
  \fix@cev{#1}{-}%
}
\newcommand{\fix@cev}[2]{%
  \ifx#1\displaystyle
    \mkern#23mu
  \else
    \ifx#1\textstyle
      \mkern#23mu
    \else
      \ifx#1\scriptstyle
        \mkern#22mu
      \else
        \mkern#22mu
      \fi
    \fi
  \fi
}

\makeatletter
\newcommand{\printfnsymbol}[1]{%
  \textsuperscript{\@fnsymbol{#1}}%
}
\@addtoreset{section}{part}
\makeatother
\titleformat{\part}[display]
{\normalfont\LARGE\bfseries\centering}{}{0pt}{}

\title{AMEIR: Automatic Behavior Modeling, Interaction Exploration \\ and MLP Investigation in the Recommender System}

\author{
Pengyu Zhao\thanks{Equal Contribution}$^{1,2}$
\and
Kecheng Xiao\printfnsymbol{1}$^1$ \and
Yuanxing Zhang\printfnsymbol{1}$^3$ \and
Kaigui Bian$^1$ \And
Wei Yan$^1$
\affiliations
$^1$School of EECS, Peking University, Beijing, China \\
$^2$HULU LLC, Beijing, China \\
$^3$Alibaba Group, Beijing, China
\emails
\{pengyuzhao, kecheng, bkg, w\}@pku.edu.cn, yuanxing.zyx@alibaba-inc.com
}

\begin{document}

\maketitle

\begin{abstract}
    Recently, deep learning models have been widely spread in the industrial recommender systems and boosted the recommendation quality. Though having achieved remarkable success, the design of task-aware recommender systems usually requires manual feature engineering and architecture engineering from domain experts.
    To relieve those human efforts, we explore the potential of neural architecture search (NAS) and introduce AMEIR for \textbf{A}utomatic behavior \textbf{M}odeling, interaction \textbf{E}xploration and multi-layer perceptron (MLP) \textbf{I}nvestigation in the \textbf{R}ecommender system. 
    The core contributions of AMEIR are the three-stage search space and the tailored three-step searching pipeline.
    Specifically, AMEIR divides the complete recommendation models into three stages of behavior modeling, interaction exploration, MLP aggregation, and introduces a novel search space containing three tailored subspaces that cover most of the existing methods and thus allow for searching better models. 
    To find the ideal architecture efficiently and effectively, AMEIR realizes the one-shot random search in recommendation progressively on the three stages and assembles the search results as the final outcome.
    Further analysis reveals that AMEIR's search space could cover most of the representative recommendation models, which demonstrates the universality of our design.
    The extensive experiments over various scenarios reveal that AMEIR outperforms competitive baselines of elaborate manual design and leading algorithmic complex NAS methods with lower model complexity and comparable time cost, indicating efficacy, efficiency and robustness of the proposed method.
\end{abstract}

\section{Introduction}

Recommender system has become an essential service on online E-commerce business and content platforms to deliver items that best fit users' interests from the substantial number of candidates.
Recently, advancements of deep learning (DL) have innovated the recommending strategies in real-world applications~\cite{covington2016deep,zhou2018deep} to enable accurate and personalized recommendation.

DL-based recommender systems have to process both sequential and non-sequential input features, and thereby follow a characterized architecture compared to the general DL tasks.
A canonical DL-based recommendation model usually includes three parts:
(1) \textbf{Behavior modeling} \cite{hidasi2018recurrent,kang2018self,tang2018personalized,zhou2018deep} probes into capturing user's diverse and dynamic interest from historical behaviors;
(2) \textbf{Interaction exploration} \cite{guo2017deepfm,lian2018xdeepfm,qu2018product,wang2017deep}
finds useful interactions among different fields to provide memorization and intuitive conjunction evidence for the recommendation.
(3) \textbf{Multi-layer perceptron (MLP) investigation} \cite{covington2016deep,zhou2018deep} aggregates the inputs with the results from behavior modeling and interaction exploration, and then fuses the features with hidden layers in the MLP.

Industry demands efficiently designing the three parts in the DL-based recommendation model to improve the accuracy and personalization for given recommender tasks in business.
Nevertheless, it is hard to find a unified model meeting the requirements in all the scenarios.
For the models of \textbf{behavior modeling}, 
recurrent neural network (RNN)~\cite{hidasi2018recurrent,hidasi2016session} is hard to preserve long-term behavioral dependencies even though employing gated memory cells.
Convolutional neural network (CNN)~\cite{tang2018personalized,yuan2019simple} is capable of learning local feature combinations yet it relies on a wide kernel, deep network or global pooling to distinguish long-term interests.
The attention mechanism~\cite{zhou2018deep} directly aggregates the entire behavior sequence, but it can not capture the evolution of the user's preference~\cite{zhou2019deep}.
Self-attention~\cite{vaswani2017attention} is better for modeling long-term behaviors~\cite{feng2019deep,sun2019bert4rec}, but it is hard to be deployed in real-time applications that require fast inference with limited resources. 
Regarding the methods for \textbf{interaction exploration} and \textbf{MLP investigation}, vanilla MLP~\cite{covington2016deep} implicitly generalizes high-order information in the network whereas missing the low-order information.
Therefore, linear regression (LR)~\cite{cheng2016wide}, factorization machine (FM)~\cite{guo2017deepfm,he2017neural2} and product-based neural network (PNN)~\cite{qu2016product,qu2018product} are introduced to involve low-order feature interactions from raw input categorical features.
On top of that, DCN~\cite{wang2017deep} and xDeepFM~\cite{lian2018xdeepfm} further learn compressed high-order interactions in the networks.
However, these methods either require hand-crafted cross features or simply enumerate interactions of bounded degree, and thus introducing noise in the model~\cite{liu2020autofis}.
Moreover, the MLP specified manually or by grid search is usually sub-optimal.
Therefore, adopting and coordinating the task-aware architectures for the \textbf{three parts} of recommendation models are the main challenge for building \textbf{efficient} and \textbf{accurate} recommender systems.

Recently, the neural architecture search (NAS) paradigm~\cite{zoph2018learning,liu2019darts,xie2019snas,real2019regularized,tan2019efficientnet,so2019evolved,wang2020textnas} 
is proposed to automatically design deep learning models by searching task-specific and data-oriented optimal architecture from the search space, thereby mitigating a lot of human efforts.
Some latest attempts~\cite{luo2019autocross,liu2020autofis,song2020towards} incorporate NAS with the recommendation.
Nevertheless, they all neglect \textbf{behavior modeling} and only consider \textbf{interaction exploration} and \textbf{MLP investigation} in the \textbf{restricted} search spaces.
Different from these methods, we introduce \textbf{AMEIR}, namely \textbf{A}utomatic behavior \textbf{M}odeling, interaction \textbf{E}xploration and MLP \textbf{I}nvestigation in the \textbf{R}ecommender system, to \textbf{automatically} find the \textbf{complete} recommendation models for adapting \textbf{various} recommender tasks and mitigating manual efforts.
Specifically, AMEIR divides the backbone architecture of recommendation models based on the three stages of behavior modeling, interaction exploration, MLP aggregation, and introduces a novel search space containing three tailored subspaces that are designed to cover most of the representative recommender systems and thus allow for searching better architectures.
Facing the industrial demands of agile development and architecture iteration, AMEIR realizes the efficient while competitive \textbf{one-shot random search}~\cite{li2019random} in this paper, and proposes a three-step searching pipeline to \textbf{progressively} find the ideal architectures for the corresponding three stages in the recommendation model.
The search results will be assembled as the final outcome.
AMEIR would not add complicated or redundant operations to the computation graph, and thus the yielded models can be optimized by the common industrial accelerating strategies~\cite{nvidia2020hugectr}.
The experimental results over various recommendation scenarios show that our automatic paradigms consistently achieves state-of-the-art performance despite the influence of multiple runs, and could outperform both the strongest baselines with elaborate manual design and leading NAS methods with comparable search cost, indicating efficacy, efficiency, and robustness of the proposed method.
Further analysis reveals that the overall time cost of AMEIR is similar to the baseline methods without architecture search and the searched models always possess lower model complexity, which demonstrates that AMEIR is efficient in both search and inference phases.

\begin{figure*}[t]
    \centering
    \includegraphics[width=\linewidth]{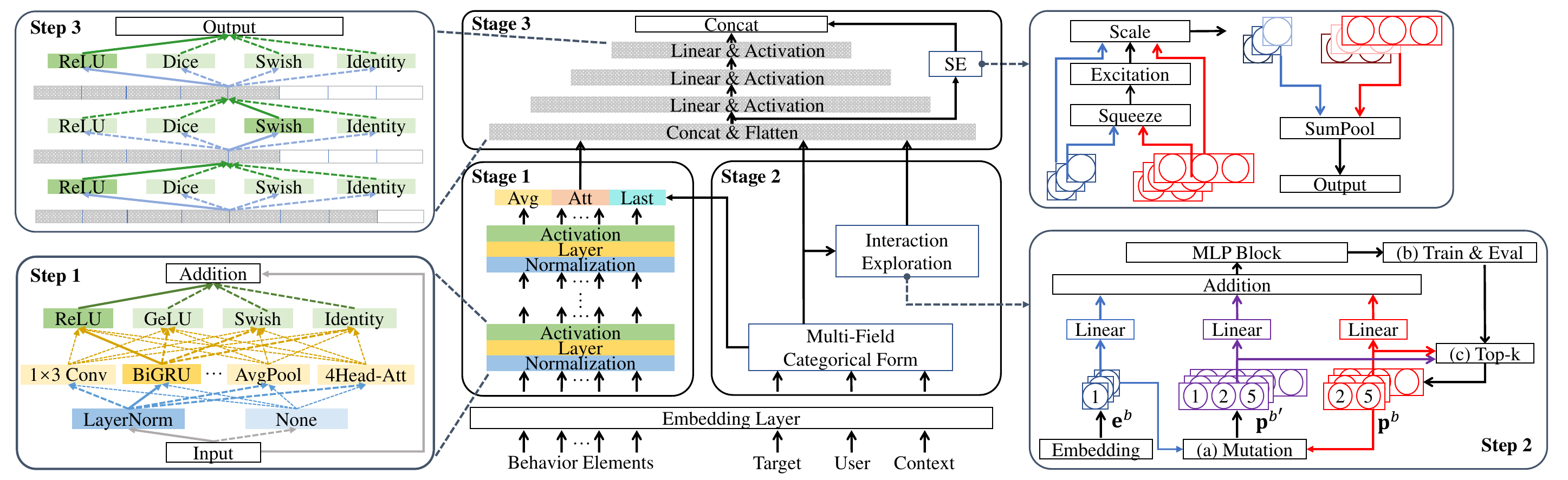}
    \caption{The three-stage search space and the three-step one-shot searching pipeline of AMEIR. AMEIR searches for building blocks, feature interactions, and aggregation MLP (three layers as an example) from corresponding search subspaces in Steps~1-3.
    }
    \label{fig:AMEIR}
\end{figure*}

\section{AMEIR}
\label{sec:AMEIR}
In Sec.\ref{sec:AMEIR}, we will introduce the three-stage search space, and the tailored three-step one-shot searching pipeline in AMEIR.
We will also discuss the relationship between AMEIR and representative recommender systems and show that AMEIR is able to cover most of these methods.

\subsection{Backbone Model in AMEIR's Search Space}
The recommendation models in AMEIR's search space share the same three-stage backbone, as illustrated in Fig.\ref{fig:AMEIR}.

\noindent\textbf{Stage~0: Feature Representation and Embedding}. 
In the common recommender systems, the input features are collected in multi-field categorical form, and
Generally, there are three basic elements in the data instances: user profile, item profile and context profile.
User profile represents the user's personal information and can be expressed by a tuple of multiple fields, e.g., ("User ID", "User Attributes", etc).
Analogously, item profile and context profile can be represented by ("Item ID", "Item Attributes", etc) and ("Time", "Place", etc).
Based on the basic elements, AMEIR segregates the features into two groups:
segregated into sequential features and non-sequential features:
(1) The sequential features describe the user's historical behaviors, which can be represented by a sequence of behavior elements.
Each behavior element contains the corresponding item profile and the optional context profile.
(2) The non-sequential features depict the attribute information of the current recommendation, including user profile, instant context profile, and the optional target profile for click-through rate (CTR) prediction task.

Conforming to~\cite{covington2016deep,zhou2018deep}, AMEIR transforms the data instances into high-dimensional sparse vectors via one-hot or multi-hot encoding.
Specifically, if the field is univalent, it will be converted to one-hot vector; if the field is multivalent, it will be encoded as a multi-hot vector.
Each data instance can be formally illustrated as:
\begin{equation}\label{eqn:0_feature_field}
    \mathbf{x} = [\mathbf{x}^{a}, \mathbf{x}^{b}] = 
    [\mathbf{x}_1^a, \mathbf{x}_2^a, ..., \mathbf{x}_T^a, \mathbf{x}_1^{b}, \mathbf{x}_2^{b}, ..., \mathbf{x}_N^{b}],
\end{equation}
where $\mathbf{x}^{a} = [\mathbf{x}_1^a, \mathbf{x}_2^a, ..., \mathbf{x}_T^a]$ denotes the encoded sequential behavior elements of length $T$ with each behavior element $\mathbf{x}_i^{a} = [\mathbf{x}_{i,1}^{a}, \mathbf{x}_{i,2}^{a}, ..., \mathbf{x}_{i,N_a}^{a}]$ grouped in $N_a$ categorical fields of the same dimension, and $\mathbf{x}^{b} = [\mathbf{x}_1^b, \mathbf{x}_2^b, ..., \mathbf{x}_N^b]$ represents $N$ encoded vectors of non-sequential feature fields.
AMEIR compresses the sparse features into low dimensional dense vectors via embedding technique. 
Suppose there are $F$ unique features.
AMEIR creates the embedding matrix $\mathbf{E} \in \mathbb{R}^{F \times K}$, where each embedding vector has dimension $K$.
The embedding process follows the embedding lookup mechanism:
If the field is univalent, the field embedding is the feature embedding;
If the field is multivalent, the field embedding is the sum of feature embedding. 
The embedded data instance can then be represented by:
\begin{equation}\label{eqn:0_feature_embedding}
    \mathbf{e} = [\mathbf{e}^a, \mathbf{e}^b] =
    [\mathbf{e}_1^a, \mathbf{e}_2^a, ..., \mathbf{e}_T^a, \mathbf{e}_1^{b}, \mathbf{e}_2^{b}, ..., \mathbf{e}_N^{b}],
\end{equation}
where $\mathbf{e}_i^{a} = [\mathbf{e}_{i,1}^{a}, \mathbf{e}_{i,2}^{a}, ..., \mathbf{e}_{i,N_a}^{a}] \in \mathbb{R}^{N_a K}$ is the concatenation of the embeddings in $\mathbf{x}_i^{a} = [\mathbf{x}_{i,1}^{a}, \mathbf{x}_{i,2}^{a}, ..., \mathbf{x}_{i,N_a}^{a}]$, and $\mathbf{e}_i^b \in \mathbb{R}^K$ is the embedding of $\mathbf{x}_i^b$.

\noindent\textbf{Stage~1: Behavior Modeling on Sequential Features}.
To extract the sequential representation from user's behavior, AMEIR introduces a \textbf{block-wise behavior modeling network} to model sequential patterns based on grouped behavioral embedding 
$\mathbf{H}_0^a = [\mathbf{e}_1^{a}, \mathbf{e}_2^{a}, ..., \mathbf{e}_T^{a}]$.
The behavior modeling network is constructed by stacking a fixed number of $L_a$ residual \textbf{building blocks}, where the $l$-th block receives the hidden state $\mathbf{H}_{l-1}^a$ from the previous block as input and generates a new hidden state $\mathbf{H}_l^a$ of the same dimension. 
Each block consists of three consecutive operations of normalization, layer, and activation, which are selected from the corresponding \textbf{operation sets} in the \textbf{search subspace of building blocks}.
The hidden state of layer $l$ is formulated as:
\begin{equation}\label{eqn:1_sequential}
    \mathbf{H}_l^a = \text{Act}_l^a(\text{Layer}_l^a(\text{Norm}_l^a(\mathbf{H}_{l-1}^a))) + \mathbf{H}_{l-1}^a, 
\end{equation}
where $\text{Act}_l^a$, $\text{Layer}_l^a$ and $\text{Norm}_l^a$ are the selected activation, layer and normalization operations for layer $l$.
As the fully connected networks can only handle fixed-length inputs, AMEIR compresses the final hidden states into fixed dimensions via sequence-level sum pooling:  $\mathbf{h}^{a}_{pool} = \sum_{i=1}^T \mathbf{h}_{L_a, i}^a$,
where $\mathbf{h}_{L_a, i}^a$ denotes the $i$-th element in $\mathbf{H}_{L_a}^a$.
Moreover, to extract the latest interest of the user, the last hidden state $\mathbf{h}_{L_a, T}^a$ is also included in the result.
Besides, when the target item is involved, AMEIR adopts an attention layer to extract a target-aware representation: $\mathbf{h}^{a}_{att} = \sum_{i=1}^T a_i \mathbf{h}_{L_a, i}^a$, with $a_i$ computed by the local activation unit
~\cite{zhou2018deep}.
The concatenation $\mathbf{h}^a = [\mathbf{h}^{a}_{pool}, \mathbf{h}_{L_a, T}^a, \mathbf{h}^{a}_{att}]$ is served as the sequential feature extracted from Stage~1.
In practice, this stage remains the procedure to execute looking up embedding only once, which is orthogonal to the industrial large-scale sparse-features lookup optimizations~\cite{oldridgemerlin}.

\noindent\textbf{Stage~2: Interaction Exploration on Non-sequential Features}.
Learning effective feature interactions is crucial for recommender systems.
Different from the widely-used methods that manually extract useful interactions or exhaustively enumerate interactions of bounded degrees~\cite{cheng2016wide,guo2017deepfm,lian2018xdeepfm}, AMEIR explores a small number of $M$ beneficial interactions $\mathbf{p}^b = [\mathbf{p}_1^b, \mathbf{p}_2^b, ..., \mathbf{p}_M^b]$ from the \textbf{search subspace of interactions}.
The search subspace contains all low-order and high-order interactions among the non-sequential features, where each interaction has the same dimension as the input embeddings, i.e., $\mathbf{p}_i^b \in \mathbb{R}^K$, $i \in [1, 2, ..., M]$.

\noindent\textbf{Stage~3: Aggregation MLP Investigation with Shortcut Connection}.
Given the inputs of sequential features, non-sequential features, and explored interactions 
$\mathbf{h}_0^c = [\mathbf{h}^a, \mathbf{e}^b, \mathbf{p}^b]$,
AMEIR jointly learns an MLP with dense \textbf{Squeeze-and-Excitation} (SE)~\cite{hu2018squeeze} connection to aggregate these features by taking advantages of memorization from wide shortcut connection and generalization from deep neural network. 
Assume the aggregation has $L_c$ hidden layers.
Considering both recommendation quality and efficiency, AMEIR searches for the \textbf{dimension} and \textbf{activation function} for each hidden layer from the \textbf{search subspace of MLP}.
The forward pass of each hidden layer $l \in [1, 2, ..., L_c]$ can be depicted as:
\begin{equation}\label{eqn:3_combination_mlp}
    \mathbf{h}_l^c = \text{Act}_l^c(\mathbf{W}_l^c \mathbf{h}_{l-1}^c + \mathbf{b}_l^c),
\end{equation}
where $\mathbf{h}_l^c$, $\mathbf{W}_l^c$, $\mathbf{b}_l^c$ and $\text{Act}_l^c$ are the output, weight, bias and activation function of the $l$-th layer.
Meanwhile, inspired by the bypaths in~\cite{cheng2016wide,guo2017deepfm,lian2018xdeepfm}, 
AMEIR introduces a dense connection to help memorize the low-order information from the feature embeddings and explored interactions $\mathbf{Q}^{se} = [\mathbf{e}^b, \mathbf{p}^b] = [\mathbf{e}^b_1, ..., \mathbf{e}^b_N, \mathbf{p}^b_1, ..., \mathbf{p}^b_M]$.
As the dimension of $\mathbf{Q}^{se}$ is several times higher than $\mathbf{h}_{L_c}^c$, AMEIR employs the SE shortcut connection with 1.0 expansion ratio for attentive recalibration and feature compression~\cite{liu2020autofis}.
The attention units $\mathbf{a}^{se} = [a^{se}_1, a^{se}_2, ..., a^{se}_{M+N}]$ in the SE network are computed by:
\begin{equation}\label{3_se_weight}
    \mathbf{a}^{se} = \sigma(\mathbf{W}^{se}_2 \text{ReLU}(\mathbf{W}^{se}_1 \mathbf{Q}^{se} \mathbf{v}^{se})),
\end{equation}
where $\mathbf{v}^{se}$ is applied for linear projection, and $\mathbf{W}^{se}_1, \mathbf{W}^{se}_2$ are weight matrices in the fully-connected layers.
The output of SE connection is given by the weighted sum $\mathbf{h}^{se} = \sum_{i=1}^{M+N} a^{se}_i \mathbf{q}_i^{se}$,
where $\mathbf{q}_i^{se}$ is the $i$-th element in $\mathbf{Q}^{se}$. 
AMEIR concatenates the results from aggregation MLP and SE connection as the output representation $\mathbf{h}^c = [\mathbf{h}_{L_c}^c, \mathbf{h}^{se}]$.

\noindent\textbf{Loss Function}.
For the CTR prediction task, AMEIR minimizes 
the binary log loss for optimization:
\begin{equation}\label{eqn:loss_1}
    L = -\frac{1}{|\mathcal{S}|} \sum_{(\mathbf{x}, y) \in \mathcal{S}} [y \log \sigma(\mathbf{w}^\mathsf{T} \mathbf{h}^c) + (1-y) \log (1 - \sigma(\mathbf{w}^\mathsf{T} \mathbf{h}^c))],
\end{equation}
where $\sigma(\mathbf{w}^\mathsf{T} \mathbf{h}^c)$ is the predicted CTR score, and $\mathcal{S}$ is the training set with $\mathbf{x}$ as input and $y \in \{0, 1\}$ as label.
For the item retrieval task, AMEIR adopts the sampled binary cross entropy loss as the objective function:
\begin{equation}\label{eqn:loss_2}
    L = -\frac{1}{|\mathcal{S}|} \sum_{(\mathbf{x}, i) \in \mathcal{S}} [\log \sigma(\mathbf{e}_{i}^\mathsf{T} \mathbf{W} \mathbf{h}^c) + \sum_{j \in \mathcal{I}_{i}^-} \log (1 - \sigma(\mathbf{e}_j^\mathsf{T} \mathbf{W} \mathbf{h}^c))],
\end{equation} 
where $\sigma(\mathbf{e}_{i}^\mathsf{T} \mathbf{W} \mathbf{h}^c)$ is the relevance score between target item $i$ and final representation; $\mathcal{S}$ is the training set of input feature $\mathbf{x}$ and target item $i$;
$\mathcal{I}_{i}^-$ is the item set of negative samples.

\subsection{Search Subspaces of Three Stages}
For each stage in the backbone, AMEIR introduces a corresponding search subspace to cover the representative recommendation models.
We will describe each of the three subspaces below and show a detailed comparison in Sec.\ref{sec:related}.

\noindent\textbf{Search Subspace of Building Blocks}.
In order to cover the existing behavior modeling methods, AMEIR proposes a \textbf{search subspace of building blocks} in the behavior modeling network, including three operation sets for normalization, layer, and activation.
Specifically, AMEIR collects the \textbf{normalization set} of \{Layer normalization~\cite{ba2016layer}, None\} and \textbf{activation set} of \{ReLU, GeLU~\cite{vaswani2017attention}, Swish~\cite{ramachandran2017searching}, Identity\} that are commonly used in the previous methods.
Regarding the layer operations, AMEIR introduces four categories of candidate layers, namely \textbf{convolutional layers}, \textbf{recurrent layers}, \textbf{pooling layers}, and \textbf{attention layers} to identify sequential patterns in the user history.
The convolutional layers are all one-dimension, including standard convolution with kernel size \{1, 3\} and dilated convolution with kernel size \{3, 5, 7\}. 
The pooling layers consist of average pooling and max pooling with kernel size 3.
Both convolutional and pooling layers are of stride 1 and SAME padding.
Bi-directional GRU (Bi-GRU) is employed as the recurrent layer as it is faster than Bi-LSTM without loss of precision. Two-head and four-head bi-directional self attentions~\cite{vaswani2017attention} are also introduced for better behavior modeling.
Additionally, when the target item appears in the input, the layer operation set will associate an attention layer that attends to each sequence position from target~\cite{zhou2018deep}. 
Besides, \textbf{zero operation} is also included to implicitly allow a dynamic depth of behavior modeling network by dropping redundant layers.
It is worth emphasizing that the operation sets are proposed to cover the \textbf{existing} methods. 
When a new operation is developed, it can be easily added to the corresponding operation set to help find a better recommendation model.

\noindent\textbf{Search Subspace of Interactions}.
The existing literature adopts \textbf{Hadamard product}~\cite{lian2018xdeepfm}, \textbf{inner product}~\cite{guo2017deepfm,qu2018product}, \textbf{bilinear function}~\cite{huang2019fibinet} and \textbf{cross product}~\cite{cheng2016wide} in the feature interactions.
Hadamard product calculates the element-wise product among the feature embeddings, which can be exploited by both MLP and shortcut connection.
The inner product can be seen as a simple form of Hadamard product compressed by sum pooling.
Bilinear function extends Hadamard product by introducing extra parameters to learn fine-grained representations, but it can not handle high-order interactions and does not exhibit superiority over Hadamard product when combining with MLP~\cite{huang2019fibinet}.
Although the cross product can yield a more flexible form of interactions, it can only be used in the shortcuts~\cite{luo2019autocross} because the parameters will exponentially explode in the order of interactions and can not be well trained due to the low frequency of occurrence.
Therefore, AMEIR chooses \textbf{Hadamard product} as the interaction function due to its expressive power and flexibility.
The \textbf{search subspace of interactions} includes all low-order and high-order Hadamard interactions.
An interaction of order $r$ can be expressed as $\mathbf{e}_{i_1}^b \odot \mathbf{e}_{i_2}^b \odot ... \odot \mathbf{e}_{i_r}^b$
where $\odot$ refers to the Hadamard product and $\mathbf{e}_{i_1}^b, \mathbf{e}_{i_2}^b, ..., \mathbf{e}_{i_r}^b$ are selected from $\mathbf{e}^b$.

\noindent\textbf{Search Subspace of MLP}.
The \textbf{search subspace of MLP} includes \textbf{dimensions} and \textbf{activation functions} of hidden layers. 
Assume $K_0^c$ is the dimension of $\mathbf{h}_0^c$.
The dimensions of hidden layers are chosen from $\{0.1, 0.2, ..., 1.0\}$ of $K_0^c$ while ensuring monotonically non-increasing following the practice.
The activation functions are selected from \{ReLU, Swish, Identity, Dice~\cite{zhou2018deep}\}.

\subsection{Three-step One-shot Searching Pipeline}
\noindent\textbf{One-shot NAS in AMEIR}.
Recent approaches~\cite{bender2018understanding,pham2018efficient} introduce efficient \textbf{one-shot weight-sharing} paradigm in NAS to boost search efficiency, where all \textbf{child models} share the weights of common operations in a large \textbf{one-shot model} that subsumes every possible architecture in the search space.
Though diverse complex search algorithms~\cite{cai2019proxylessnas,chen2019detnas,chu2019fairnas,dong2019network} have been coupled with one-shot NAS, 
the latest studies~\cite{bender2018understanding,dong2020bench,li2019random} reveal that the efficient \textbf{one-shot random search} is surprisingly better than the complex gradient-based methods of DARTS~\cite{liu2019darts}, SNAS~\cite{xie2019snas} and RL-based methods of ENAS~\cite{pham2018efficient}.
To facilitate the industrial demands of agile development and architecture iteration, AMEIR realizes the one-shot random search in the recommendation model:
(1) The child models are \textbf{randomly} sampled from the search spaces (actually subspaces) to train the shared weights of the one-shot model and validate its performance on a small subset of validation instances to ensure adequate training while avoiding overfitting.
(2) A subset of child models is \textbf{randomly} selected and evaluated on the same validation subset using the inherited weights from the one-shot model.
(3) The child models with the best performance will be derived and retrained for the final recommendation outcome. 
Moreover, to reduce the magnitude of the search space and find the better recommendation models, AMEIR modularizes the search process into \textbf{three} steps matching the three stages in the backbone model by progressively searching for the architectures in the three search subspaces\cite{li2019improving,sciuto2020evaluating}. 
Meanwhile, AMEIR ensures the fairness~\cite{chu2019fairnas} of one-shot model training to alleviate the representation shift caused by weight sharing~\cite{chu2019fairnas,yu2020train}, where all candidates are uniformly sampled and activated per gradient update.
Details of the three steps are shown below.

\noindent\textbf{Step~1: Behavior Modeling Search}.
In Step~1, AMEIR searches for the behavior modeling networks from the search subspace of building blocks, as shown in Fig.\ref{fig:AMEIR}. 
The normalization, layer, and activation operations are randomly selected from the corresponding operation sets to construct the recommendation model.
To decouple behavior modeling from other stages, a pre-defined MLP is exploited to combine $\mathbf{h}^a$ and $\mathbf{e}^b$ during Step~1.
The training and evaluation procedures are the same as previously described.

\noindent\textbf{Step~2: Interaction Exploration}.
In Step~2, AMEIR combines one-shot NAS with the sequential model-based optimization (SMBO) algorithm~\cite{liu2018progressive} to explore interactions on non-sequential features.
The interaction set $\mathbf{p}^b$ is initialized with non-sequential features $\mathbf{e}^b$, and then progressively updated by several cycles of evolution to increase the interaction order, as illustrated in Fig.\ref{fig:AMEIR}:
(a) The candidate interactions in $\mathbf{p}^b$ intersect with all non-sequential fields from $\mathbf{e}^b$ via Hadamard product to mutate for higher-order interactions. The expanded interaction set is denoted by $\mathbf{p}^{b'}$.
(b) The intersections in $\mathbf{p}^{b'}$ are trained by one-shot NAS and evaluated on the validation set.
(c) Top-$k$ interactions with the highest validation fitness are retained as the new $\mathbf{p}^b$, while other interactions are filtered out to avoid exhaustively visiting all possible solutions.
After the evolution, AMEIR selects top-$M$ features in $\mathbf{p}^b$ as the resulting interaction set.

In order to train and evaluate interactions in the one-shot model, AMEIR utilizes another pre-defined MLP to aggregate interactions in $\mathbf{p}^{b'}$ with non-sequential embeddings in $\mathbf{e}^b$, while the behavior modeling here is left out for efficiency.
As the linear projection of feature concatenation is equivalent to feature-wise addition of projected results, i.e., 
$\mathbf{W} \mathbf{h}^\mathsf{T} = [\mathbf{W}_1, \mathbf{W}_2, ..., \mathbf{W}_N] [\mathbf{h}_1, \mathbf{h}_2, ..., \mathbf{h}_N]^\mathsf{T} = \sum_{i=1}^N \mathbf{W}_{i} \mathbf{h}_{i}^{\mathsf{T}}$,
the one-shot NAS in Step~2 is efficiently realized by adding the projections of one uniformly-sampled interaction and all embeddings in $\mathbf{e}^b$ as the output of the first hidden layer, without accessing to all interactions in $\mathbf{p}^{b'}$.
To further speed up the search process, the weights of the one-shot model are inherited in consecutive evolution cycles.

\noindent\textbf{Step~3: MLP Investigation}.
In Step~3, AMEIR investigates the dimensions and activation functions in the aggregation MLP based on the behavior modeling network and interactions found in the previous steps.
To find the hidden sizes of MLP, AMEIR pre-allocates a weight matrix with the maximum dimension of ($K_0^c$, $K_0^c$), i.e.,  $\mathbf{W}^c \in \mathbb{R}^{K_0 \times K_0}$, for each hidden layer~\cite{guo2019single}.
Suppose $h_{\text{in}}$ and $h_{\text{out}}$ are chosen to be the hidden sizes of previous and current layers. 
AMEIR will slice out sub-matrix $\mathbf{W}^c[:h_{\text{in}}, :h_{\text{out}}]$ to assemble the MLP in Eqn.~(\ref{eqn:3_combination_mlp}).
Other training and evaluation procedures are similar to Steps~1 and 2.

\noindent \textbf{Derivation}.
After the three-step random search, AMEIR assembles the search results as the final outcome and retrains the derived model for evaluation.
It is worth emphasizing that the derived model is initialized with the one-shot model weights because the training protocols of model architectures and datasets are the same for both one-shot model and child models, which is in contrast to the most existing methods that adopt proxy task for one-shot model training~\cite{liu2019darts,so2019evolved,xie2019snas,yu2020train}.  
The inherited weights accelerate the retraining process and compensate for the search cost.

\subsection{Related Works to AMEIR}
\label{sec:related}

\noindent\textbf{Behavior Modeling}. 
The existing methods introduce RNN~\cite{hidasi2018recurrent}, CNN~\cite{yuan2019simple}, or Transformer-based~\cite{sun2019bert4rec} behavior modeling networks 
to capture sequential relations and evolved interests from user behaviors.
Then, the sequential features are extracted from the sequence-level hidden states by either taking the last hidden state~\cite{kang2018self} or aggregating the sequence by a pooling layer~\cite{cheng2016wide,covington2016deep}.
Some recent works~\cite{zhou2018deep,zhou2019deep} further employ an attention mechanism between the sequential representations and target item to retrieve the user's main proposal.
These methods can all be found in AMEIR's search space.
Specifically, the search subspace of building blocks contains all layer, activation, and normalization operations in those behavior modeling networks.
Moreover, the last hidden state, pooling-based aggregation, and attention-based aggregation are all used for sequence compression.

\noindent\textbf{Interaction Exploration}.
In parallel with behavior modeling, some works resort to explore feature interactions in the DL-based recommendation models. 
LR~\cite{cheng2016wide,luo2019autocross}, FM~\cite{guo2017deepfm,liu2020autofis}, DCN~\cite{wang2017deep} and xDeepFM~\cite{lian2018xdeepfm} are exploited to import explicit low-order or high-order interactions through shortcut connections, i.e., concatenating interactions to the last hidden layer of MLP, and then apply linear projection with learnable variables or all-one vectors to generate the prediction scores jointly with the hidden layers.
Other than the shortcut connection, FibiNet~\cite{huang2019fibinet} take interactions as the input of MLP, while PNN~\cite{qu2016product,qu2018product} further combines interactions with raw embeddings to induce more features in the network.
It is not hard to find that all these methods can be derived from AMEIR's search space.
Concretely, all low-order and high-order interactions can be extracted from the search subspace of interactions.
Meanwhile, raw input embeddings and explored interactions are used by both shortcut connection and MLP, covering the existing interaction-combining strategies.
Besides, the attentive SE connection could implicitly characterize attention-based, regression-based, and FM-based shortcut connections. 

\begin{table}[htp]
\centering
\caption{Comparison with representative methods on Alimama.}
\resizebox{\linewidth}{!}{
\begin{tabular}{c|c|c|c}
\toprule
Category & Model & AUC & QPS\\
\midrule
\multirow{3}{*}{DNN-based} & DNN~\cite{covington2016deep} & 0.6304 & 64K \\
& Wide\&Deep~\cite{cheng2016wide} & 0.6313 & 55K \\
& DeepFM~\cite{guo2017deepfm} & 0.6320  & 58K \\
\midrule
\multirow{3}{*}{Attention-based} & DIN~\cite{zhou2018deep} & 0.6341 & 53K \\
& DIEN~\cite{zhou2019deep} & 0.6344 & 44K\\
& DSIN~\cite{feng2019deep} & 0.6352 & 43K \\
\midrule
\multirow{3}{*}{AMEIR} & AMEIR-A & 0.6357 $\pm$ 0.0001 & 47K \\
& AMEIR-B & 0.6359 $\pm$ 0.0003 & 43K \\
& AMEIR-C & \textbf{0.6383} $\pm$ 0.0004 & 46K \\
\bottomrule
\end{tabular}
}
\label{tab:hybrid}
\end{table}

\begin{table}[htp]
\centering
\caption{Comparison with representative methods on Beauty/Steam.}
\resizebox{\linewidth}{!}{
\begin{tabular}{c|c|c|c|c}
\toprule
\multirow{2}{*}{Model} & \multicolumn{2}{c|}{Beauty} & \multicolumn{2}{c}{Steam} \\
\cmidrule{2-5}
 & HR@5 & NDCG@5 & HR@5 & NDCG@5 \\
\midrule
POP & 0.175 & 0.114 & 0.517 & 0.365 \\
BPR~\cite{rendle2009bpr} & 0.301 & 0.227 & 0.527 & 0.369 \\
GRU4REC$^+$~\cite{hidasi2018recurrent} & 0.300 & 0.215 & 0.626 & 0.468\\
STAMP~\cite{liu2018stamp} & 0.215 & 0.149 & 0.642 & 0.483 \\
NARM~\cite{li2017neural} & 0.243 & 0.166 & 0.658 & 0.501 \\
Caser~\cite{tang2018personalized} & 0.298 & 0.217 & 0.670 & 0.510 \\
NextItNet~\cite{yuan2019simple} & 0.334 & 0.243 & 0.679 & \textbf{0.517} \\
SRGNN~\cite{wu2019session} & 0.214 & 0.148 & 0.651 & 0.494 \\
BERT4REC~\cite{sun2019bert4rec} & 0.337 & 0.252 & 0.676 & 0.513\\
\midrule
AMEIR-A & 
 \tabincell{c}{\textbf{0.341}\\$\pm$ 0.004} & \tabincell{c}{\textbf{0.257}\\$\pm$ 0.003} &  \tabincell{c}{\textbf{0.681}\\$\pm$ 0.003} & \tabincell{c}{\textbf{0.517}\\$\pm$ 0.002} \\
\bottomrule
\end{tabular}
}
\label{tab:sequential}
\end{table}

\noindent \textbf{MLP Investigation}.
The existing literature~\cite{covington2016deep,guo2017deepfm,zhou2018deep} mainly uses sub-optimal hand-crafted MLPs to aggregate features, where activation functions and hidden sizes are designed manually or by grid search.
However, the search space of MLP is large and these hand-crafted MLP settings are always sub-optimal.
AMEIR introduces a discrete yet flexible search space that covers most of the common MLP settings and finds the MLP settings through the reliable and learnable neural architecture search.

Based on the above analysis, it can be inferred that AMEIR's search space covers most of the representative recommendation models.
Therefore, AMEIR can perform as good as these methods in various scenarios, and further derive better models when conducting sufficient training and exhaustive search.
A more detailed comparison between AMEIR and the related works is given in Appendix~\ref{sec:app_A}.

\section{Experiments}
\label{sec:exp}
In Sec.\ref{sec:exp}, we will compare AMEIR with competitive baselines on various recommendation scenarios, study the impact of different components, and conduct the efficiency analysis on the proposed method.
We will also describe more details of the experiments in Appendix~\ref{sec:app_B}-\ref{sec:app_E}, including datasets, baseline methods, implementations, and search results.

\subsection{Industrial Results for AMEIR}
To verify all three steps of AMEIR in the industrial scenario, we refer to the Alimama CTR dataset~\cite{feng2019deep} that comprises both sequential and non-sequential features for a comprehensive comparison.
We use similar preprocessing settings as \cite{feng2019deep}, where the embedding size is set to 4 and the maximum sequence length is set to 50.
The Adam optimizer with an initial learning rate 1e-5 and batch size 1024 is employed for both AMEIR and baseline methods.
For the training of the one-shot model, a single cosine schedule is introduced for learning rate decay.
In Steps~1 and~3, AMEIR selects the top-5 models from the randomly sampled 2000 architectures for further evaluation, while in Step~2, AMEIR conducts 4 cycles of evolution, retains top-50 interactions at each cycle, and finally reserves $M=5$ interactions as $\mathbf{p}^b$.
$L_a$ and $L_c$ are set to 3 and 2 in the backbone model, and the pre-defined MLP is set to [200, 80] during Steps~1 and~2.
By conforming to~\cite{li2019random,liu2019darts}, we run AMEIR for 4 times and report the mean metrics across 4 runs as the final results.

Tab.~\ref{tab:hybrid} depicts the results on Alimama dataset. 
AMEIR-A (only uses behavior modeling result from Step~1) presents competitive performance with the promising empirical architectures DIEN and DSIN.
Furthermore, a significant improvement (note that an improvement of AUC around 0.0005-0.001 is already regarded as practically significant in the industrial scenarios~\cite{feng2019deep}) can be observed when combining AMEIR-A with searched interactions (AMEIR-B) and MLP (AMEIR-C).
An interesting point on the Alimama dataset is that compared to sequential modeling and interaction exploration, the MLP search significantly contributes to the final results, which
indicates that all three stages/steps in AMEIR are necessary for high-quality recommendations.

\begin{table}[htp]
\centering
\caption{Comparison with representative methods on Criteo/Avazu.}
\resizebox{\linewidth}{!}{
\begin{tabular}{c|c|c|c|c}
\toprule
\multirow{2}{*}{Model} & \multicolumn{2}{c|}{Criteo} & \multicolumn{2}{c}{Avazu} \\
\cmidrule{2-5}
 & AUC & Log Loss & AUC & Log Loss \\
\midrule
DNN~\cite{covington2016deep} & 0.7983 & 0.4549 & 0.7746 & 0.3831 \\
Wide \& Deep~\cite{cheng2016wide} &  0.7992 & 0.4538 & 0.7749 & 0.3839 \\
DeepFM~\cite{guo2017deepfm} & 0.801 & 0.4496 & 0.7753 & 0.3825  \\
IPNN~\cite{qu2018product} & 0.7975 & 0.4578  &  0.7787 & 0.3806 \\
DCN~\cite{wang2017deep} & 0.7981 & 0.4611 & 0.7772 & 0.3815 \\
xDeepFM~\cite{lian2018xdeepfm}  & 0.8019 & 0.4497  & 0.7759 & 0.3818 \\
FibiNet~\cite{huang2019fibinet}  & 0.8021  & 0.4487 & 0.7789 & 0.3798  \\
\midrule
AutoCross~\cite{luo2019autocross} & 0.8022 $\pm$ 0.0001 & 0.4494 $\pm$ 0.0001 & 0.7776 $\pm$ 0.0002 & 0.3810 $\pm$ 0.0001 \\
AutoFIS~\cite{liu2020autofis}  & 0.8015 $\pm$ 0.0004 & 0.4504 $\pm$ 0.0005 & 0.7789 $\pm$ 0.0003 & 0.3801 $\pm$ 0.0002 \\
AutoCTR~\cite{song2020towards}  & 0.8027 $\pm$ 0.0003 & 0.4489 $\pm$ 0.0003  & 0.7791 $\pm$ 0.0002 & 0.3800 $\pm$ 0.0001 \\
\midrule
AMEIR-I & 0.8024 $\pm$ 0.0002  & 0.4491 $\pm$ 0.0004  & 0.7791 $\pm$ 0.0003  & 0.3799 $\pm$ 0.0003 \\
AMEIR-B  & 0.8030 $\pm$ 0.0002 & 0.4486 $\pm$ 0.0004   & 0.7798 $\pm$ 0.0004 & 0.3794 $\pm$ 0.0002 \\
AMEIR-C  & \textbf{0.8034}  $\pm$ 0.0002 & \textbf{0.4482} $\pm$ 0.0003   & \textbf{0.7802} $\pm$ 0.0004 & \textbf{0.3792} $\pm$ 0.0002 \\
\bottomrule
\end{tabular}
}
\label{tab:non-sequential}
\end{table}

\begin{table}[tp]
\centering
\caption{Comparison with NAS methods on Alimama.}
\resizebox{\linewidth}{!}{
\begin{tabular}{c|c|c}
\toprule
Model & AUC & Search Cost (GPU days)\\
\midrule
Random & 0.6321 $\pm$ 0.0011 & 2.9 \\
RL~\cite{zoph2017neural} & 0.6323 $\pm$ 0.0006 & 3.2 \\
EA~\cite{real2019regularized} & 0.6338 $\pm$ 0.0004 & 3 \\
\midrule
ENAS (One-shot + RL)~\cite{pham2018efficient} & 0.6339 $\pm$ 0.0007 & 0.15 \\
DARTS (One-shot + Gradient)~\cite{liu2019darts} & 0.6350 $\pm$ 0.0008 & 0.2 \\
SNAS (One-shot + Gradient)~\cite{xie2019snas} & 0.6343 $\pm$ 0.0003 & 0.3\\
\midrule
One-shot Random Search~\cite{li2019random} & 0.6365 $\pm$ 0.0006 & 0.3 \\
AMEIR (Three-stage One-shot Random Search) & \textbf{0.6383} $\pm$ 0.0004 & 0.5\\
\bottomrule
\end{tabular}
}
\label{tab:search}
\end{table}

\begin{table}[tp]
\centering
\caption{Comparison of backbone models on Criteo/Avazu.}
\resizebox{\linewidth}{!}{
\centering
\begin{tabular}{p{1.7cm}<{\centering} | p{1.2cm}<{\centering}|p{1cm}<{\centering}|p{1.4cm}<{\centering}|p{1cm}<{\centering}|p{1.4cm}<{\centering}}
 \toprule
 \multicolumn{2}{c|}{Model} &
 \multicolumn{2}{c|}{Criteo} & \multicolumn{2}{c}{Avazu} \\ \midrule
 Interaction & MLP & AUC & Log Loss & AUC &  Log Loss \\ 
 \midrule
 None & Manual & 0.7983 & 0.4549 & 0.7746 & 0.3831 \\ 
 MLP & Manual & 0.8021 & 0.4492 & 0.7789 & 0.3801 \\ 
 FM & Manual & 0.8024 & 0.4491 & 0.7791 & 0.3799 \\ 
 MLP + FM & Manual  & 0.8027 & 0.4488 & 0.7795 & 0.3795 \\
 MLP + SE & Manual & 0.8030 & 0.4486 & 0.7798 & 0.3794 \\ 
  \midrule
 MLP + SE & Searched & \textbf{0.8034} & \textbf{0.4482} & \textbf{0.7802} & \textbf{0.3792}\\
 \bottomrule
 \end{tabular}
}

\label{tab:ablation}
\end{table}

\subsection{Applicable for Various Scenarios}
To show AMEIR is applicable for various recommendation tasks, we evaluate AMEIR on two additional scenarios, namely sequential scenario, and non-sequential scenario:
(1) The sequential scenario only presents sequential user behaviors to examine Step~1 behavior modeling.
The training is based on the item retrieval task.
The sequential datasets include Amazon Beauty and Steam~\cite{kang2018self}.
(2) The non-sequential scenario is introduced to verify the Step~2 interaction exploration and Step~3 MLP investigation, where only non-sequential features are provided for the recommendation.
The non-sequential datasets include benchmarking CTR datasets Criteo~\cite{guo2017deepfm} and Avazu~\cite{qu2018product}.
The implementation details are similar to the industrial scenario and can be found in Appendix~\ref{sec:app_D}.

\noindent \textbf{Evaluation on Sequential Scenario}.
Tab.~\ref{tab:sequential} summarizes the results on Beauty and Steam datasets.
The hand-crafted methods, including the strongest baselines BERT4Rec and NextItNet, can not dominate both datasets, which empirically justifies that the best architecture is usually task-aware.
In contrast, 
AMEIR consistently achieves better accuracy compared to the baselines with well-designed architectures, indicating the strength of AMEIR and the capability of searching better models from the search subspace of building blocks.

\noindent \textbf{Evaluation on Non-sequential Scenario}.
Tab.~\ref{tab:non-sequential} presents the results on non-sequential datasets.
It can be observed that AMEIR-I (DeepFM backbone with interactions explored by AMEIR) consistently surpasses competitive hand-crafted baselines xDeepFM and FibiNet as well as AutoML-based baselines AutoCross and AutoFIS on different tasks over multiple runs, validating the efficacy of interaction exploration in Step~2.
Moreover, our method gains remarkable improvement over AMEIR-I and exceeds AutoCTR when combining explored interactions with the SE connection (AMEIR-B) and MLP investigation (AMEIR-C).
The result verifies the design of both the backbone and search subspaces.

\subsection{Ablation Study}
\noindent \textbf{Ablation Study on NAS Methods}.
To study the importance of the proposed search method, we evaluate the performance of different NAS algorithms on the same search space of the Alimama dataset. 
As shown in Tab.~\ref{tab:search}, the one-shot random search achieves better performance than both brute-force methods of random, RL, EA, and the algorithmic complex methods of ENAS, DARTS, SNAS with lower search cost.
Moreover, AMEIR with the three-stage one-shot searching pipeline brings about a remarkable improvement on the one-shot random search under a similar time budget, demonstrating the effectiveness and efficiency of the proposed method.

\noindent \textbf{Ablation Study on Backbone Model}.
To further analyze the impact of the backbone model, we isolate different components in Stages~2 and~3, and report the mean metrics on the non-sequential datasets.
As shown in Tab.~\ref{tab:ablation}, the interactions (searched) applied to MLP and FM bring orthogonal improvement on vanilla DNN.
The result reveals that both combining strategies contribute to the result.
Additionally, SE interaction (AMEIR-B) and MLP search (AMEIR-C) could further boost the performance of recommendation, suggesting the superiority of the backbone design in AMEIR.

\subsection{Efficiency in Search and Inference Phases}
\noindent \textbf{Search Efficiency}.
The search cost of AMEIR on the Beauty, Steam, Avazu, Criteo, Alimama datasets are 0.2, 0.2, 0.25, 0.25, 0.5 GPU-days respectively on a single Tesla V100 GPU, which are all similar to training the individual models of baseline methods.
As AMEIR uses the trained one-shot weights to initialize the derived model, the fine-tuning time could generally be ignored.
Therefore, the overall time cost of AMEIR is roughly the same as the methods that do not employ NAS. 
The result proves the efficiency of AMEIR in the search phase, and it is evident that AMEIR can help facilitate the design of recommender systems in real-world scenarios.

\noindent \textbf{Inference Efficiency}. 
We record the queries/samples-per-second (QPS) of AMEIR and the compared models to examine the inference speed on the Alimama dataset in Tab.~\ref{tab:search}.
Apart from the improvement in accuracy, AMEIR is also competitive in efficiency compared with the commonly-used attention-based models.
Besides, the searched MLPs are always much smaller than the hand-crafted ones (details shown in Appendix~\ref{sec:app_E}), such that the numbers of parameters of the searched models are less than half of the baselines.
The results indicate that instead of increasing the model complexity, AMEIR coordinates the three parts of a recommender system to find the architecture with the best practice.

\section{Conclusion}

In this paper, we introduce AMEIR for automatic behavior modeling, interaction exploration, and MLP investigation in the recommender systems to relieve the manual efforts of feature engineering and architecture engineering.
Owing to the three-stage search space and the matched three-step searching pipeline, AMEIR covers most of the representative recommendation models and outperforms the competitive baselines in various recommendation scenarios with lower model complexity and comparable time cost, demonstrating efficiency, effectiveness, and robustness of the proposed method.

\bibliographystyle{named}
\bibliography{AutoRec}

\newpage
\onecolumn

\part{Appendix}

\renewcommand\thesection{\Alph{section}}

\section{Detailed Comparison between AMEIR and Representative Methods} \label{sec:app_A}

In Appendix~\ref{sec:app_A}, we conduct a detailed comparison between AMEIR and the representative recommender systems on behavior modeling and interaction exploration to show that these methods are all covered by AMEIR's search space.

\subsection{Behavior Modeling}

The existing behavior modeling methods can be categorized into RNN-based, CNN-based, Transformer-based methods, and mixed methods.
\begin{itemize}
    \item RNN-based methods~\cite{hidasi2016session,hidasi2018recurrent,quadrana2017personalizing,wu2017recurrent} mainly apply LSTM or GRU layers to extract sequential representations, which can be covered by the building blocks with recurrent layer operations.
    \item CNN-based methods apply convolutional layers on the sequential representations.
    Caser~\cite{tang2018personalized} introduces horizontal and vertical convolutions by sliding windows on the sequential embedding and applies max pooling over the extracted feature maps for compression.
    However, sequence-level pooling on one hidden convolution layer is hard to capture complex relations in the sequence. 
    NextItNet~\cite{yuan2019simple} utilizes stacked residual blocks of dilated convolution, which can be roughly regarded as a special case of the behavior modeling network in AMEIR because both dilated convolutional block and standard $1 \times 1$ convolutional block are included in the search subspace of Stage~1.
    \item Transformer-based methods, e.g., ATRank~\cite{zhou2018atrank}, SASRec~\cite{kang2018self}, and BERT4Rec~\cite{sun2019bert4rec} introduce self-attention layers and point-wise feed-forward layers to capture long-term semantics, which are the same as stacking multi-head attention blocks and $1 \times 1$ convolutional blocks in the behavior modeling network.
    \item Mixed methods such as DIEN~\cite{zhou2019deep} and DSIN~\cite{feng2019deep} employ the mixture of the above-mentioned layers, which potentially take advantages of these operations from various aspects.
    DIEN introduces GRU + AUGRU to capture interest evolution in user behavior.
    DSIN employs self-attention + Bi-LSTM to model inner- and intra-relationship in multiple historical sessions.
    Both DIEN and DSIN (if neglecting the session division) can be approximated by sequences of building blocks in AMEIR with the specified layer operations, i.e., \{Bi-GRU, Attention from Target, Bi-GRU\} and \{4-Head Self Attention, $1 \times 1$ Convolution, Bi-GRU\} respectively.
\end{itemize}

On top of the behavior modeling, these methods either extract the last hidden state~\cite{hidasi2016session,kang2018self}, or employ pooling layer~\cite{tang2018personalized} and attention layer from target~\cite{zhou2018deep,zhou2019deep} to compress the sequential hidden states into a fixed-length representation, which are all comprised by the behavior modeling module of Stage~1 in AMEIR.

\subsection{Interaction Exploration}

In Sec.2, we have declared that AMEIR can cover all feature interactions and interaction-combining strategies in the representative interaction exploration methods.
For a more formal and intuitive comparison, we develop the computation process of final prediction scores in both AMEIR and the representative recommender systems and show that all compared methods can be explicitly or implicitly included in AMEIR's search space.
Following the existing symbol usage, $\mathbf{x}^b$ and $\mathbf{e}^b$ denote the non-sequential input features and corresponding field embeddings.
For the convenience of representation, we use $\mathbf{p}^b$ to represent both investigated interactions and enumerated bounded-degree interactions of various types, including cross-product interactions, bilinear interactions, and Hadamard-product interactions (inner-product interaction can be expressed as a special form of Hadamard-product interaction, i.e., $\mathbf{1}^\mathsf{T} \mathbf{p}_i^b$, where $\mathbf{p}_i^b$ represents a Hadamard-product interaction) according to the related literature.
We use $\mathbf{a}$ to denote the attention weights of interactions and embeddings computed by compressed interaction network (CIN)~\cite{lian2018xdeepfm} or attention module~\cite{liu2020autofis,xiao2017attentional}.
The bias in the MLP layers is omitted for convenience.
We compare AMEIR with representative methods of LR, FM~\cite{rendle2010factorization}, AFM~\cite{xiao2017attentional}, DNN~\cite{covington2016deep}, Wide \& Deep~\cite{cheng2016wide}, AutoCross~\cite{luo2019autocross} (Wide \& Deep version), NFM~\cite{he2017neural2}, DeepFM~\cite{guo2017deepfm}, DCN~\cite{wang2017deep}, xDeepFM~\cite{lian2018xdeepfm}, AutoFIS~\cite{liu2020autofis}, PNN~\cite{qu2018product} and FibitNet~\cite{huang2019fibinet} (SE feature interactions implicitly included in $\mathbf{p}^b$):
\begin{align}
    \mathbf{y}_{\text{LR}} &= \mathbf{w}_x^\mathsf{T} \mathbf{x}^b \\
    \mathbf{y}_{\text{FM}} &= \mathbf{1}^\mathsf{T} \sum_i \mathbf{p}_i^b + \mathbf{w}_x^\mathsf{T} \mathbf{x}^b\\
    \mathbf{y}_{\text{AFM}} &= \mathbf{w}_p^\mathsf{T} (\sum_i a_i \mathbf{p}_i^b) + \mathbf{w}_x^\mathsf{T} \mathbf{x}^b \\
    \mathbf{y}_{\text{DNN}} &= \mathbf{w}_M^\mathsf{T} \text{MLP}(\mathbf{e}^b) \\
    \mathbf{y}_{\text{Wide \& Deep}} &= \mathbf{w}_M^\mathsf{T} \text{MLP}(\mathbf{e}^b) + \mathbf{w}_p^\mathsf{T} \mathbf{p}^b + \mathbf{w}_x^\mathsf{T} \mathbf{x}^b \\
    \mathbf{y}_{\text{AutoCross}} &= \mathbf{w}_M^\mathsf{T} \text{MLP}(\mathbf{e}^b) + \mathbf{w}_p^\mathsf{T} \mathbf{p}^b + \mathbf{w}_x^\mathsf{T} \mathbf{x}^b \\
    \mathbf{y}_{\text{NFM}} &= \mathbf{w}_M^\mathsf{T} \text{MLP}(\mathbf{e}^b, \sum_i \mathbf{p}_i^b) + \mathbf{w}_x^\mathsf{T} \mathbf{x}^b \\
    \mathbf{y}_{\text{DeepFM}} &= \mathbf{w}_M^\mathsf{T} \text{MLP}(\mathbf{e}^b) + \mathbf{1}^\mathsf{T} \sum_i \mathbf{p}_i^b + \mathbf{w}_x^\mathsf{T} \mathbf{x}^b \\
    \mathbf{y}_{\text{DCN}} &= \mathbf{w}_M^\mathsf{T} \text{MLP}(\mathbf{e}^b) + \mathbf{w}_p^\mathsf{T} (\sum_{i} a_i \mathbf{p}_i^b) + \mathbf{w}_e^\mathsf{T} (\sum_{j} \mathbf{e}_j^b)  \\
    \mathbf{y}_{\text{xDeepFM}} &= \mathbf{w}_M^\mathsf{T} \text{MLP}(\mathbf{e}^b) + \mathbf{w}_p^\mathsf{T} (\sum_{i} a_i \mathbf{p}_i^b) +  \mathbf{w}_x^\mathsf{T} \mathbf{x}^b \\
    \mathbf{y}_{\text{AutoFIS}} &= \mathbf{w}_M^\mathsf{T} \text{MLP}(\mathbf{e}^b) + \mathbf{1}^\mathsf{T} (\sum_i a_i \mathbf{p}_i^b) + \mathbf{w}_x^\mathsf{T} \mathbf{x}^b \\
    \mathbf{y}_{\text{FibiNet}} &= \mathbf{w}_M^\mathsf{T} \text{MLP}(\mathbf{p}^b) + \mathbf{w}_x^\mathsf{T} \mathbf{x}^b\\
    \mathbf{y}_{\text{PNN}} &= \mathbf{w}_M^\mathsf{T} \text{MLP}(\mathbf{e}^b, \mathbf{p}^b) \\
    \mathbf{y}_{\text{AMEIR}} &= \mathbf{w}_M^\mathsf{T} \text{MLP}(\mathbf{e}^b, \mathbf{p}^b) + \mathbf{w}_p^\mathsf{T} (\sum_{i} a_i \mathbf{p}_i^b) +  \mathbf{w}_e^\mathsf{T} (\sum_{j} a_j \mathbf{e}_j^b),
\end{align}
where $\mathbf{w}_x$, $\mathbf{w}_e$, $\mathbf{w}_p$ and $\mathbf{w}_M$ are the weights of linear projections on the input features, input embeddings, feature interactions, and final hidden state of MLP. 
It can be inferred that the MLP modules $\text{MLP}(\mathbf{e}^b)$ (MLP on embedding), $\text{MLP}(\mathbf{p}^b)$ (MLP on interactions) and $\text{MLP}(\mathbf{e}^b, \sum_i \mathbf{p}_i^b)$ (MLP on embedding and compressed interactions) are all special cases of $\text{MLP}(\mathbf{e}^b, \mathbf{p}^b)$ (MLP on embedding and interactions), while the interaction modules $\mathbf{1}^\mathsf{T} \sum_i \mathbf{p}_i^b$ (FM on interactions), $\mathbf{w}_p^\mathsf{T} \mathbf{p}^b$ (regression on interactions) and $\mathbf{1}^\mathsf{T} (\sum_i a_i \mathbf{p}_i^b)$ (FM on weighted interactions) are all comprised by SE connection $\mathbf{w}_p^\mathsf{T} (\sum_{i} a_i \mathbf{p}_i^b)$ (regression on weighted interactions).
Besides, the bypath $\mathbf{w}_e^\mathsf{T} (\sum_{j} a_j \mathbf{e}_j^b)$ can implicitly capture the semantics of linear regression on the raw inputs of $\mathbf{w}_x^\mathsf{T} \mathbf{x}^b$ and cross network on the input embedding of $\mathbf{w}_e^\mathsf{T} (\sum_{j} \mathbf{e}_j^b)$.
Therefore, all compared methods are explicitly or implicitly contained in AMEIR's search space.

\section{Datasets} \label{sec:app_B}
In Appendix~\ref{sec:app_B}, we introduce the datasets in the industrial scenario, sequential scenario, and non-sequential scenario.
Tables~\ref{tab:hybrid_stat}\--\ref{tab:nonseq_stat} show the statistics of the datasets.

\subsection{Industrial Scenario}

\begin{table}[h]
\centering
\caption{Statistics of Alimama dataset in the industrial scenario.}
\resizebox{\linewidth}{!}{
\begin{tabular}{cccccccc}
 \toprule
 \# Users & \# Sparse Feature & \# Dense Feature & \# (cate\_id, brand) & Avg. Seq. Length & \# Train Traces & \# Test Traces & Prop. of Click \\
 \midrule
 265,443 & 15 & 1 & 1,027,443 & 46.31 & 5,544,213 & 660,694 & 0.0514 \\
 \bottomrule
 \end{tabular}
}
\label{tab:hybrid_stat}
\end{table}

The Alimama dataset in the industrial scenario contains data of all formats mentioned in Sec.~3 so that it is used for comprehensive comparison in all three-stage pipeline of AMEIR:
Alimama$\footnote{https://tianchi.aliyun.com/dataset/dataDetail?dataId=56}$ provides 26 million records of ad display/click logs in 8 days, where for each user it keeps track of 200 recent shopping behaviors in 22 days.
The logs in the first 7 days are used for training and the logs in the last day are treated as the test set.
Each data instance has 15 sparse features, including user profile, user's sequential behaviors, and context profile.
The dimension of these features ranges from 3 to 309,449.
There is only 1 dense feature of ''price'' which has been normalized to [-1,1].
We randomly sample 25\% of the dataset and remains 6,204,907 data instances following~\cite{feng2019deep}.
5\% of the training set is used for validation in the architecture search and model selection.
As there are many users that do not have any records of sequential behaviors, we assign a special tag to the behavior sequences of these users to avoid the behavior modeling network executing on vacant inputs.

\subsection{Sequential Scenario}

\begin{table}[h]
\centering
\caption{Statistics of Beauty and Steam datasets in the sequential scenario.}
\begin{tabular}{ccccccc}
 \toprule
& \# User & \# Item & Min. Length & Max. Length & Avg. Length & Density\\
\midrule
Beauty & 22,363 & 12,101 & 5 & 204 & 8.876 & 0.07\%\\
Steam & 281,210 & 11,961 & 5 & 1,226 &  12.391 & 0.10\%\\
 \bottomrule
 \end{tabular}
\label{tab:seq_stat}
\end{table}

The Amazon Beauty and Steam datasets in the sequential scenario are proposed to evaluate the Stage~1 behavior modeling in AMEIR, which only include sequential behavior of each user, and the training is based on next-item prediction with Eqn.~(7):
\begin{itemize}
    \item Amazon$\footnote{http://jmcauley.ucsd.edu/data/amazon/}$ is composed of reviews and metadata from Amazon.com~\cite{mcauley2015image}, where data instances are categorized into separate datasets according to the product category. In this paper, we use the category of ``Beauty'' for evaluation.
    \item Steam$\footnote{https://cseweb.ucsd.edu/\textasciitilde jmcauley/datasets.html\#steam\_data}$ is introduced by~\cite{kang2018self} which contains millions of review information from the Steam video game platform. 
\end{itemize}
Following the common practice~\cite{he2017neural,kang2018self}, the reviews in the datasets are treated as implicit feedback.
We group data instances by users and sort them according to the timestamps.
The users and items with less than 5 feedbacks are removed from datasets.
The leave-one-out strategy is adopted for partitioning~\cite{he2017neural,kang2018self,sun2019bert4rec}, where for each user the last two actions are treated as the validation set and test set, and the remaining actions are used for training.

\subsection{Non-sequential Scenario}

\begin{table}[h]
\centering
\caption{Statistics of Criteo and Avazu datasets in the non-sequential scenario.}
\begin{tabular}{ccccccc}
 \toprule
& \# Train Instances & \# Test Instances & Fields & Prop. of Click\\
\midrule
Criteo & 41,256,555
 & 4,584,062
 & 39 & 25.62\%\\
Avazu & 32,343,173
 & 8,085,794
 & 24 & 16.98\%\\
 \bottomrule
 \end{tabular}
\label{tab:nonseq_stat}
\end{table}

The Criteo and Avazu datasets in the non-sequential scenario only include non-sequential categorical features for CTR prediction to verify the Stage~2 interaction exploration and Stage~3 aggregation MLP investigation:
\begin{itemize}
    \item Criteo$\footnote{http://labs.criteo.com/downloads/download-terabyte-click-logs/}$ contains 7 days of click-through data, which is widely used for CTR prediction benchmarking. There are 26 anonymous categorical fields and 13 continuous fields in the Criteo dataset.
    The continuous fields are normalized to [-1,1] and only used for MLP input while categorical fields are applied to both MLP and shortcut connection as well as the interaction exploration module in AMEIR.
    The dataset is randomly split into two parts while maintaining the label distribution: 90\% is for training and the rest is for testing. 
    Besides, 10\% of the training set is used for validation.
    \item Avazu$\footnote{http://www.kaggle.com/c/avazu-ctr-prediction}$ consists of 11 days of online ad click-through data. 
    Each data instance includes 24 categorical fields that represent profiles of a single ad impression.
    We randomly split the dataset into two parts, where 80\% is for training and the rest is for testing.
    Same as the Criteo dataset, 10\% of the training set is used for validation.
\end{itemize}

\section{Baseline Methods}  \label{sec:app_C}

In Appendix~\ref{sec:app_C}, we introduce baseline methods that are adopted for comparison in the industrial scenario, sequential scenario, non-sequential scenario, and ablation study on searching methods.

\subsection{Industrial Scenario}
The baselines in the industrial scenario include the primitive YouTube DNN~\cite{covington2016deep}, Wide \& Deep~\cite{cheng2016wide}, DeepFM~\cite{guo2017deepfm} and the more competitive methods of DIN~\cite{zhou2018deep}, DIEN~\cite{zhou2019deep}, DSIN~\cite{feng2019deep}.

\begin{itemize}
    \item YouTube DNN~\cite{covington2016deep} is usually introduced as the base model in CTR prediction.
    It utilizes sum pooling on the users’ historical behavior embedding and then concatenates the results with other non-sequential embeddings of the user profile, context profile, and target item/ad profile as the input of MLP to generate the final prediction score.
    \item Wide \& Deep~\cite{cheng2016wide} additionally adds a shortcut connection of linear regression on the non-sequential features to the output of MLP to capture both generalization and memorization.
    \item DeepFM~\cite{guo2017deepfm} replaces the linear regression-based shortcut connection in Wide \& Deep with a Factorization Machine to involve feature interactions in the recommendation model.
    \item DIN~\cite{zhou2018deep} extends the base MLP model by using an attention mechanism on the user behavior with respect to the target item to adaptively learn the diverse characteristic of user interests.
    \item DIEN~\cite{zhou2019deep} improves DIN by using GRU and AUGRU module to the extract user's evolving interests, where an auxiliary loss is employed to make hidden states more expressive. 
    \item DSIN~\cite{feng2019deep} further investigates the homogeneity and heterogeneity in the user behavior and models the user interests through intra-session self-attention layer and inter-session Bi-LSTM layer. 
    The sequential user behaviors are split into sessions by 30-minute intervals, and each session remains at most 10 behaviors.
\end{itemize}

The aggregation MLP of all baseline methods is fixed to [200, 80], while the activation function is set to Dice in DIN, DIEN, and set to ReLU in other methods. 
The training is based on the CTR prediction of Eqn.~(6) for all methods.

\subsection{Sequential Scenario}
To evaluate AMEIR in the sequential scenario, we introduce several baselines for comparison.
We carefully categorize the state-of-the-art sequential recommender systems and select the representative methods for each category to report the performance in the limited pages of the paper.
Since other sequential recommendation methods have been outperformed on the similar datasets by the selected baselines, we omit comparison against them.

\begin{itemize}
    \item POP is the simplest baseline which only recommends the most popular items with the highest number of interactions from the candidate set (also used for evaluating the complexity of datasets).
    \item BPR~\cite{rendle2009bpr} is a legacy method for matrix factorization from implicit feedback and is optimized by a pairwise ranking loss.
    We use BPR to represent the family of matrix factorization-based methods~\cite{he2017neural,mnih2008probabilistic}.
    \item GRU4Rec+~\cite{hidasi2018recurrent} is an improved version of GRU4Rec, which utilizes a delicately-designed loss function and sampling strategy to train the GRU model.
    GRU4REC+ can be seen as the representative of the RNN-based sequential recommender systems, such as GRU4REC \cite{hidasi2016session}, EvoRNN \cite{belletti2019quantifying}, etc.
    \item STAMP~\cite{liu2018stamp} simultaneously captures user's long-term general interests and short-term current interests with an attention module. STAMP is a typical sequential recommender system without RNN or CNN, representing the MF-based sequential recommender system such as FPMC~\cite{rendle2010factorizing} and TransRec~\cite{He2017Translation}.
    \item NARM~\cite{li2017neural} uses a global GRU encoder to extract sequential behavior and captures the user’s main purpose with the attention mechanism on a local GRU encoder. Regarding the local GRU encoder, the latest 3 items are used for the Beauty dataset and the latest 5 items are used for the Steam dataset. NARM is a representative for the combination of attention and recurrent neural network.
    \item Caser~\cite{tang2018personalized} is a CNN-based recommender system. It applies horizontal and vertical convolutional operations on the embedding matrix for sequential representation. For Beauty dataset, the number of channels is set to 5 in the vertical convolutions and 2 in the horizontal convolutions. For Steam dataset, the number of channels is set to 8 in the vertical convolutions and 3 in the horizontal convolutions. To ensure a fair comparison, we replace the concatenation of $d$-dimension embedding in the original design of Caser (as introduced by Eqn.~(10) in \cite{tang2018personalized}) by a summation to avoid the unfair use of the embedding with double dimension.
    \item NextItNet~\cite{yuan2019simple} employs residual blocks of dilated convolutions to increase receptive fields and model both long-range and short-range item dependencies on user behaviors. We set the kernel size of convolutions as 3, and leverage 5 dilated convolutional layers with dilation 1, 2, 1, 2, 2.
    \item SRGNN~\cite{wu2019session} applies GNN to generate item representations and then encodes global and current user interests using an attention network. We use the Gated-GNN to learn the item embeddings, add regularization of all parameters with weight 1e-6, and set the initial learning rate to 1e-7 to avoid the gradient explosion of the recurrent unit. SRGNN is the classical implementation to use GNN on recommendation, which represents the GNN-based recommender system such as GCSAN~\cite{xu2019graph}, GraphRec~\cite{fan2019graph} and PinSage~\cite{ying2018graph}.
    \item BERT4Rec~\cite{sun2019bert4rec} adopts BERT~\cite{devlin2019bert} to learn bi-directional representations of sequences by predicting masked items in the context. The transformer is set to have 2 layers with 4 heads.
    The dropout rate is set to 0.5 on inputs, 0.2 on feed-forward networks after each multi-head attention and convolution to prevent overfitting.
    BERT4REC is a successful transplant of Transformer to recommendation, which is on behalf of Transformer-based recommender systems including SASREC~\cite{kang2018self}, MBTREC~\cite{zhang2020preference}, BST~\cite{chen2019behavior}, ATRank~\cite{zhou2018atrank}, etc.
\end{itemize}

The model-relevant hyper-parameters and initializations are either following the recommended settings in the corresponding papers or tuned by the grid search on the validation sets.
We employ the same training process as AMEIR for these methods.

\subsection{Non-sequential Scenario}

The baseline methods in the non-sequential scenario include representative 
interaction exploration methods in deep learning-based recommender systems.
YouTube DNN~\cite{covington2016deep} and Wide \& Deep~\cite{cheng2016wide} are early attempts of deep learning recommendation methods.
DeepFM~\cite{guo2017deepfm}, DCN~\cite{wang2017deep} and xDeepFM~\cite{lian2018xdeepfm} enumerate interactions of bounded degrees and combine the feature interactions with MLP via shortcut connection, while IPNN~\cite{qu2016product} and FibiNet~\cite{huang2019fibinet} directly use the feature interactions as the input of MLP.
AutoCross~\cite{luo2019autocross}, AutoFIS~\cite{liu2020autofis} and AutoCTR~\cite{song2020towards} are the recently-proposed AutoML-oriented methods.
They automatically find useful cross features, FM-based interactions, or aggregation modules in the recommendation model.

\begin{itemize}
    \item YouTube DNN~\cite{covington2016deep} is the base MLP model that takes raw embeddings as the input to learn implicit interactions among the features.
    \item Wide \& Deep~\cite{cheng2016wide} extends DNN by introducing shortcut connection of linear regression on the input features to learn both high-order and low-order interactions.
    \item DeepFM~\cite{guo2017deepfm} replaces the linear regression in Wide \& Deep with the FM module.
    \item IPNN~\cite{qu2016product} combines the inner product of feature interactions with raw feature embeddings as MLP input.
    \item DCN~\cite{wang2017deep} proposes cross network to comprise explicit bit-wise interactions of bounded degrees. The compressed interactions and the output of MLP are concatenated and then fed to a linear layer to generate the prediction scores.
    \item xDeepFM~\cite{lian2018xdeepfm} designs compressed interaction network (CIN) to replace cross network in DCN~\cite{wang2017deep} by enumerating vector-wise interactions via Hadamard product. 
    The linear regression is also applied for shortcut connection as DeepFM.
    \item FibiNet~\cite{huang2019fibinet} employs SENet~\cite{hu2018squeeze} to learn the importance of embeddings and then generates bilinear interactions on both SE and original feature embeddings as the input of MLP. The linear regression is also used in FibiNet.
    \item AutoCross~\cite{luo2019autocross} learns useful high-order cross feature interactions on a tree-structured space with beam search and successive mini-batch gradient descent. The explored interactions are exploited in the linear part of Wide \& Deep. 
    \item AutoFIS~\cite{liu2020autofis} identifies important feature interactions in the FM model with differentiable architecture search~\cite{liu2019darts}, and keeps the architecture parameters as attention weights for the corresponding interactions. The generated interactions are applied to the FM module in DeepFM.
    \item AutoCTR~\cite{song2020towards} modularizes representative operations of one-layer MLP, dot product (DP) and factorization machine (FM) to formulate the direct acyclic graphs search space for the recommendation model. It performs an evolutionary algorithm with learning-to-rank guidance during architecture search. The data sub-sampling and hash size reduction are used for acceleration.
\end{itemize}

To ensure fair comparison, we use the same pre-defined aggregation MLP in AMEIR for the baseline methods, i.e., [200 $\times$ 3] for Criteo~\cite{guo2017deepfm} and [500 $\times$ 5] for Avazu~\cite{qu2018product} with ReLU activation.
Besides, CIN is set to 256 / 128 layers for Criteo / Avazu in xDeepFM.
The training is based on the CTR prediction of Eqn.~(6) for all methods.

\subsection{Ablation Study on Search Methods}

The baseline search methods include the brute-force random search, reinforcement learning (RL), evolutionary algorithm (EA), and the popular one-shot methods ENAS, DARTS, SNAS.
Conforming to the backbone settings in Sec.3, we select the best performing behavior modeling network, top-5 feature interactions, and top-1 aggregation MLP in different stages for all methods.

\begin{itemize}
    \item Random~\cite{bergstra2012random} is the baseline NAS method that simply samples architectures from a uniform distribution over the search space.
    For a fair comparison, we randomly select 300 child models from the search space until the total training time plus the evaluation time reaches 10 times the search cost of the proposed one-shot random search.
    All child models are trained by 1 epoch according to ~\cite{dong2019bench,radosavovic2020designing}, and the top-5 architectures with the highest validation accuracy will be selected for further evaluation to pick the best one as the final result.
    
    \item RL~\cite{zoph2017neural,zoph2018learning} is another baseline method of NAS.
    At each round, a controller generates a child model from the search space.
    The child model is then trained and evaluated on the validation set, and the result is treated as the reward to train the weights of the controller for a better outcome.
    In our implementation, we use a two-layer LSTM as the controller with the hidden size of 32.
    The controller is trained by REINFORCE rule~\cite{williams1992simple} with Proximal Policy Optimization~\cite{schulman2017proximal}, with a 0.9 exponential moving average of the previous rewards as the baseline~\cite{pham2018efficient}.
    We also use the entropy penalty, temperature 5, and the tanh constant 2.5 for the sampling logits~\cite{pham2018efficient}.
    Each child model is trained for 1 epoch that is the same as the random search.
    The searching process is finished once the performance of the consecutively generated models does not grow or reaches the 10-time budgets as the one-shot random search. 
    Similar to the Random, top-5 models in history will be selected for more evaluation.
    
    \item EA~\cite{real2017large,real2019regularized} maintains a population of child models, and gradually evolves the population for better results.
    In this paper, we re-implement the regularized evolutionary algorithm~\cite{real2019regularized} for comparison.
    Specifically, we first uniformly sample several child models from the search space as the initial population.
    To mutate the population, we sample one random child model as the parent from the population at each cycle, and then construct a new model from the parent by a mutation, which causes a simple modification on the network architecture, e.g, altering a layer operation in Stage~1, a feature interaction in Stage~2, or an MLP hidden size in Stage~3.
    Once the child architecture is constructed, it is then trained, evaluated on the validation set, and added to the population.
    Meanwhile, the oldest member of the population will be removed to maintain the population size.
    We set the initial population size as 10, and evolve only one architecture at each cycle.
    Each child model in the population is trained by 1 epoch.
    Once the population does not change for over 5 cycles or reach the 10-time budgets as the one-shot random search, the algorithm will be finished, and the top-performing models will be derived from the population.
    
    \item ENAS~\cite{pham2018efficient} introduces the one-shot strategy in the RL algorithm.
    The only difference between ENAS and RL is that the sampled child models in ENAS share the same weights in the one-shot model.
    
    \item DARTS~\cite{liu2019darts} extends the one-shot strategy by relaxing the discrete search space to be continuous and coupling the architecture parameters with the model parameters to determine the importance of each operation in the one-shot model.
    The architecture parameters are jointly optimized in the one-shot training along with the model parameters.
    Specifically, DARTS performs a bi-level optimization, where the model weights and architecture weights are optimized with training data and validation data alternatively.
    Conforming to the original implementation, we train the architecture parameters with the Adam optimizer and apply early stopping to avoid overfitting on the validation set.
    The other training settings of the one-shot model is the same as the one-shot random search in AMEIR.
    After the one-shot modeling training, we retain the strongest operations determined by the architecture parameters as the final result. 
    
    \item SNAS~\cite{xie2019snas} can be regarded as a variant of DARTS.
    SNAS uses a unified training loss as the optimization target and replaces the softmax function on the architecture parameters in DARTS with the Gumbel-softmax~\cite{jang2017categorical} to imitate the sampling process on the discrete architecture distribution for better consistency between the one-shot model and the derived model.
    The other settings in SNAS are the same as DARTS.
    
    \item One-shot Random Search~\cite{li2019random} is roughly the same as AMEIR, except that the three stages are searched together in a one-shot model and the 6000 randomly selected architectures are used for evaluation to choose the final outcome.
\end{itemize}

For the fair comparison, we use the same training and evaluation settings in all the search methods.

\section{Implementation Details}  \label{sec:app_D}

In Appendix~\ref{sec:app_D}, we introduce the implementation details of AMEIR and baseline methods, including the common settings and the selection of hyper-parameters in industrial, sequential, and non-sequential scenarios.

\subsection{Industrial Scenario}

For the Alimama dataset in the industrial scenario, we use the embedding size 4~\cite{feng2019deep}, batch size 1024, maximum sequence length 50, and Adam optimizer~\cite{kingma2015adam} for all methods.
The embeddings and self-attention layers are initialized uniformly from $[-0.05, 0.05]$; The convolutional block and MLP layers are initialized by He normal~\cite{Kaiming2016Delving}; The GRU block is initialized by orthogonal initialization~\cite{saxe2014exact}.
We apply grid-search to find the best choices of the hyper-parameters in AMEIR and baseline methods.
We examine the impact of initial learning rate and regularizer for all methods from the candidate sets \{1e-3, 1e-4, 1e-5, 1e-6\} and \{1e-3, 1e-4, 1e-5, 1e-6\} respectively.
Meanwhile, we attempt to add dropout to both building blocks in the behavior modeling network and hidden layers in the aggregation MLP from \{0, 0.1, 0.2, 0.3, 0.4, 0.5\}.

\subsection{Sequential Scenario}

For the sequential scenario, we fix the maximum sequence length as 15 for Beauty and 25 for Steam.
The initialization is the same as industrial scenario, and the models are optimized by Adam optimization with batch size 512.
The input sequences are randomly clipped for the model training, with the last five items served as the positive items and another five randomly sampled items that the user has not interacted with served as the negative items.
We also apply grid-search for the hyper-parameter tuning in the sequential scenario.
We search the embedding size from \{64, 96, 128, 160, 192\} and find that when the embedding size is greater than 128, the performance of most of the baselines would not further increase and the compared models even become hard to reach convergence.
Therefore, we fix the embedding size as 128 for all models.
We also search for the initial learning rate from \{1e-3, 1e-4, 1e-5\}, L2 regularizer from \{1e-3, 1e-4, 1e-5, 1e-6\}, and dropout ratio of building blocks from \{0, 0.1, 0.2, 0.3, 0.4, 0.5\}.

\subsection{Non-sequential Scenario}

For the non-sequential scenario, we use an embedding size of 5 for Criteo and 40 for Avazu.
The batch size is set to 4096 and the initialization is the same as the industrial scenario.
Similar as the candidate sets in the industrial scenario and sequential scenario, learning rate, L2 regularizer and dropout ratio of MLP layers are chosen from \{1e-3, 1e-4, 3e-5, 1e-5\}, \{1e-3, 1e-4, 1e-5, 1e-6\} and \{0, 0.1, 0.2, 0.3, 0.4, 0.5\} respectively.

\section{Search Results}  \label{sec:app_E}

In Appendix~\ref{sec:app_E}, we show the searched building blocks, feature interactions, and aggregation MLPs of AMEIR and the corresponding experimental results in the 4 runs.

\subsection{Industrial Scenario}

Tab.~\ref{tab:hybrid_search} demonstrates the searched architectures on the industrial dataset.
Regarding step~1, AMEIR automatically raises the best behavior modeling network with two or three building blocks based on the sequential data.
It can be observed that the searched blocks over the 4 runs all include one non-parameterized operation of average pooling or dummy layer of zero, indicating that the shallow sequential models might be optimal for this dataset.
Therefore, compared to the empirical state-of-the-art CTR prediction models such as DIEN and DSIN, the searched building blocks contribute more to the AUC metric.
The search results also demonstrate that the one-shot training is capable of finding architectures of appropriate depth to prevent pre-mature convergence of the shallow models and avoid overfitting of the deep models.
Besides, although the ''attention from target'' operation is involved in the industrial scenario, we find that it does not help improve the AUC result.
Regarding step~2, the searched interactions contain both second-order and third-order interactions.
Though there are few non-sequential sparse features in the Alimama dataset, the searched interactions can slightly but consistently improve the performance upon the searched blocks.
Note that some interactions, e.g., (F04, F10), (F05, F06, F11), and (F06, F11) are shared across different runs.
The phenomenon verifies the robustness of AMEIR and enhances the interpretability of the prediction.
Regarding step~3, the searched aggregation MLP tends to select the layers with lower dimension activated by ReLU.
This decision significantly upgrades the performance on the CTR prediction task, with fewer parameters compared to the models by manual architecture engineering.
The prominent performance of the low-dimension MLP enlightens us that a proper MLP could boost the learning capability of a specific architecture, and one-shot searching is able to acquire the best settings of the MLP.
Additionally, compared to the novel activations of Swish and Dice, AMEIR prefers ReLU in the searched MLP, which demonstrates that using all ReLUs as the activation functions might be a good starting point for MLP design.
The above observations verify the practicability and efficiency of both the three-stage search space and the three-step one-shot searching pipeline of AMEIR.

\subsection{Sequential Scenario}

Tab.~\ref{tab:seq_search} summarizes the searched building blocks in the sequential scenario with the corresponding experimental results.
Similar to the industrial scenario, though the maximal number of layers is set to 6, all searched architectures for each dataset contains three to five layers (i.e., at least one ``Zero'' in each searched architecture), indicating that the shallow models might be more suitable for Beauty and Steam datasets.
It is also evident that the one-shot search tends to raise the self-attention block and convolutional block in those two datasets and construct similar architectures to the state-of-the-art manual designs (e.g., \#1 of Beauty to NextItNet, \#1 of Steam to BERT4REC).
Beyond that, the layer normalization is usually deployed on the shallow layers, and the Bi-GRU layer mostly locates at the deep layer.
We believe that these placements contribute to the efficient knowledge extraction from user behaviors as well as the fast convergence of sequential modeling.

\subsection{Non-sequential Scenario}

Tab.~\ref{tab:nonseq} depicts the searched interactions, as well as the pre-defined / searched MLP of AMEIR-B and AMEIR-C in the non-sequential scenario.
Though AMEIR could search high-order interactions, most of the searched interactions (best-performing interactions) are in second-order while some pairs of features never form appropriate interactions to be selected by the one-shot model according to the statistics.
This observation attributes to the automatic recognition of the feasible interactions with appropriate order by the one-shot searching, potentially relieving the effort of manual feature engineering.
Same as the industrial scenario, the shared interactions can also be found in the non-sequential scenario, e.g., (C6, C13), (C13, C26) in Criteo and (C15, C18), (device\_conn\_type, C18) in Avazu, once again verifies the robustness of AMEIR over multiple runs.
Regarding the aggregation MLP, we notice that AMEIR-C tends to select MLPs of the higher dimension in the bottom layers and lower dimension in the top layers, which is more reasonable for the neural network design as opposed to the hand-crafted settings in \cite{guo2017deepfm,qu2018product}.
The adaptive design of MLP further improves the performance on the non-sequential datasets, indicating the effective and efficient one-shot searching of AMEIR.
Besides, the preference of ReLU also appears in the non-sequential scenario.

\section{Discussion on One-shot Strategy}  \label{sec:app_F}

The one-shot strategy has been widely used for accelerating the neural architecture search.
In most of the existing literature~\cite{liu2019darts,xie2019snas,wu2019fbnet,guo2019single}, the one-shot NAS is only applied to derive the best model from the search space, and the derived model is re-trained from scratch.
In AMEIR, we find an interesting phenomenon that fine-tuning the derived model with the one-shot model weights can achieve better performance than directly re-training the derived model with random initialization.
The phenomenon is aligned with the recently proposed Once-for-All~\cite{cai2019once}, BigNAS~\cite{yu2020bignas} and AttentiveNAS~\cite{wang2020attentivenas}, which only trains the network once and directly use the one-shot weights for evaluation without re-training.
These methods achieve similar or even better performance than training the child models from scratch, which validates our experimental results.
Moreover, we observe a larger improvement in using one-shot weights as the initialization in AMEIR.
We guess that this is because the recommendation models usually encounter the overfitting problem, 
and the one-shot training, which is equivalent to a adding large drop-path strategy on the one-shot model, can be regarded as introducing an extra regularization on the model training.
Therefore, it helps prevent/delay the appearance of overfitting meanwhile guaranteeing sufficient training of the network.
We observe that under similar training settings, one-shot training presents a later overfitting epoch compared to re-training the derived models from scratch, which is consistent with our argument.

\begin{table}[tp]
\centering
\caption{Searched building blocks, interactions, MLPs and corresponding experimental results of behavior modeling (AMEIR-A), interaction exploration (AMEIR-B) and aggregation MLP investigation (AMEIR-C) on Alimama over 4 runs. 
Each block $i$ denotes the building block of $i$-th layer, which is represented by either a tuple of (layer operation, activation operation, normalization operation) or a ''Zero'' operation (skip this layer).
Each interaction is represented by a tuple of non-sequential feature fields.
Each MLP includes a list of hidden layers, where each hidden layer is represented by a tuple of (hidden size, activation operation).}
\begin{threeparttable}
\resizebox{\linewidth}{!}{
\begin{tabular}{p{1.3cm}<{\centering}|p{6.5cm}|p{3.6cm}|p{2cm}<{\centering}|p{4cm}|p{1.3cm}<{\centering}}
 \toprule
 Run ID & Searched Blocks & Searched Interactions & Model & MLP & AUC \\
 \midrule
 \multirow{6}{*}{1} & \multirow{6}{6.3cm}{\tabincell{l}{Block 1: (1$\times$7 Dconv, Swish, None)\\Block 2: Zero \\Block 3: (1$\times$5 Dconv, Identity, LayerNorm)}} & \multirow{2}{*}{None} & \multirow{2}{*}{AMEIR-A} & \multirow{4}{3.7cm}{[(200, ReLU), (80, ReLU)]} & \multirow{2}{*}{0.6358} \\
&  & & & & \\
\cmidrule{3-4}\cmidrule{6-6}
&  & \multirow{4}{3.6cm}{[(F08, F11), (F04, F10), (F06, F11), (F07, F09, F10), (F05, F06, F11)]} & \multirow{2}{*}{AMEIR-B} & & \multirow{2}{*}{0.6362} \\
& & & & & \\
 \cmidrule{4-6}
&  & & \multirow{2}{*}{AMEIR-C} & \multirow{2}{3.7cm}{[(42, ReLU), (10, ReLU)]} & \multirow{2}{*}{0.6387} \\
&  & & & & \\
 \midrule
 \multirow{6}{*}{2} & \multirow{6}{6.3cm}{\tabincell{l}{Block 1: (1$\times$3 Dconv, Identity, None)\\Block 2: (1$\times$3 AvgPool, Swish, None) \\Block 3: (1$\times$3 Conv, Identity, None)}} & \multirow{2}{*}{None} & \multirow{2}{*}{AMEIR-A} & \multirow{4}{3.7cm}{[(200, ReLU), (80, ReLU)]} & \multirow{2}{*}{0.6357} \\
&  & & & & \\
\cmidrule{3-4}\cmidrule{6-6}
&  & \multirow{4}{3.6cm}{[(F04, F10), (F03, F06), (F06, , F07, F11), (F03, F08, F11), (F05, F06, F11)]} & \multirow{2}{*}{AMEIR-B} & & \multirow{2}{*}{0.6358} \\
& & & & & \\
 \cmidrule{4-6}
&  & & \multirow{2}{*}{AMEIR-C} & \multirow{2}{3.7cm}{[(42, ReLU), (21, ReLU)]} & \multirow{2}{*}{0.6379} \\
&  & & & & \\
 \midrule
 \multirow{6}{*}{3} & \multirow{6}{6.3cm}{\tabincell{l}{Block 1: (2-Head Self Attention, GeLU, None)\\Block 2: (Bi-GRU, ReLU, LayerNorm)\\Block 3: (1$\times$3 AvgPool, Identity, None)}} & \multirow{2}{*}{None} & \multirow{2}{*}{AMEIR-A} & \multirow{4}{3.7cm}{[(200, ReLU), (80, ReLU)]} & \multirow{2}{*}{0.6356} \\
&  & & & & \\
\cmidrule{3-4}\cmidrule{6-6}
&  & \multirow{4}{3.6cm}{[(F04, F10), (F06, F11), (F01, F06), (F05, F06, F10), (F03, F05, F07)]} & \multirow{2}{*}{AMEIR-B} & & \multirow{2}{*}{0.6357} \\
& & & & & \\
 \cmidrule{4-6}
&  & & \multirow{2}{*}{AMEIR-C} & \multirow{2}{3.7cm}{[(42, ReLU), (10, ReLU)]} & \multirow{2}{*}{0.6383} \\
&  & & & & \\
 \midrule
 \multirow{6}{*}{4} & \multirow{6}{6.3cm}{\tabincell{l}{Block 1: Zero \\Block 2: (1$\times$5 Dconv, ReLU, None)\\Block 3: (Bi-GRU, Identity, LayerNorm) }} & \multirow{2}{*}{None} & \multirow{2}{*}{AMEIR-A} & \multirow{4}{3.7cm}{[(200, ReLU), (80, ReLU)]} & \multirow{2}{*}{0.6358} \\
&  & & & & \\
\cmidrule{3-4}\cmidrule{6-6}
&  & \multirow{4}{3.6cm}{[(F08, F11), (F07, F10), (F04, F06), (F06, F10), (F06, F11)]} & \multirow{2}{*}{AMEIR-B} & & \multirow{2}{*}{0.6361} \\
& & & & & \\
 \cmidrule{4-6}
&  & & \multirow{2}{*}{AMEIR-C} & \multirow{2}{3.7cm}{[(42, ReLU), (10, ReLU)]} & \multirow{2}{*}{0.6385} \\
&  & & & & \\
 \bottomrule
 \end{tabular}
}
\begin{tablenotes}
\footnotesize
        \item[*] F01: `userid'; F02: `adgroup\_id'; F03: `pid'; F04: `cms\_segid'; F05: `cms\_group\_id'; F06: `final\_gender\_code'; F07: `age\_level'; 
        \item[ ] F08: `pvalue\_level'; F09: `shopping\_level'; F10: `occupation'; F11: `new\_user\_class\_level'; F12: `campaign\_id'; F13: `customer'; 
        \item[ ] F14: `cate\_id'; F15: `brand'.
\end{tablenotes}
\end{threeparttable}
\label{tab:hybrid_search}
\end{table}

\begin{table}[tp]
\centering
\caption{Searched building blocks and corresponding experimental results of behavior modeling (AMEIR-A) on Beauty and Steam over 4 runs. 
Each block $i$ denotes the building block of $i$-th layer, which is represented by either a tuple of (layer operation, activation operation, normalization operation) or a ''Zero'' operation (skip this layer).}
\resizebox{0.8\linewidth}{!}{
\begin{tabular}{p{1.5cm}<{\centering}|p{1.2cm}<{\centering}|p{8.5cm}|p{1.3cm}<{\centering}|p{1.3cm}<{\centering}|p{1.3cm}<{\centering}}
 \toprule
 Dataset & Run ID & Searched Blocks & HR@1 & HR@5 & NDCG@5 \\
  \midrule
  \multirow{16}{*}{\tabincell{c}{Beauty}} &  1 & \tabincell{l}{Block 1: (1$\times$5 Dconv, GeLU, LayerNorm)\\Block 2: (1$\times$5 Dconv, Identity, None)\\Block 3: Zero\\Block 4: (1$\times$7 Dconv, Identity, None)\\Block 5: Zero\\Block 6: (1$\times$7 Dconv, Identity, None)
} & 0.164 & 0.345 & 0.260 \\
  \cmidrule{2-6}
   & 2 & \tabincell{l}{Block 1: (4-Head Self Attention, Swish, LayerNorm)\\Block 2: (2-Head Self Attention, GeLU, LayerNorm)\\Block 3: (4-Head Self Attention, Identity, LayerNorm)\\Block 4: Zero\\Block 5: (Bi-GRU, Swish, LayerNorm)\\Block 6: (1$\times$5 Dconv, Swish, None)
}  & 0.153 & 0.337 & 0.254 \\
  \cmidrule{2-6}
 & 3 & \tabincell{l}{Block 1: (1$\times$3 Conv, ReLU, LayerNorm)\\Block 2: Zero\\Block 3: Zero\\Block 4: (4-Head Self Attention, Identity, LayerNorm)\\Block 5: (4-Head Self Attention, Swish, LayerNorm)\\Block 6: (1$\times$3 Dconv, Swish, None)
} & 0.159 & 0.341 & 0.255 \\
  \cmidrule{2-6}
  & 4 & \tabincell{l}{Block 1: (2-Head Self Attention, Swish, LayerNorm)\\Block 2: (4-Head Self Attention, Swish, LayerNorm)\\Block 3: (1$\times$3 Dconv, ReLU, None)\\Block 4: (Bi-GRU, Swish, LayerNorm)\\Block 5: Zero\\Block 6: (Bi-GRU, Swish, None)
} & 0.158 & 0.341 & 0.256 \\
  \midrule
  \multirow{16}{*}{Steam} & 1 & \tabincell{l}{Block 1: Zero \\Block 2: (4-Head Self Attention, Swish, LayerNorm)\\Block 3: (2-Head Self Attention, Identity, LayerNorm)\\Block 4: Zero\\Block 5: (1$\times$5 Dconv, GeLU, None)\\Block 6: (1$\times$5 Dconv, GeLU, None)
} & 0.339 & 0.682 & 0.517 \\
  \cmidrule{2-6}
   & 2 & \tabincell{l}{Block 1: (2-Head Self Attention, GeLU, LayerNorm) \\ Block 2: Zero \\Block 3: (4-Head Self Attention, Identity, LayerNorm)\\Block 4: Zero\\Block 5: Zero\\Block 6: (Bi-GRU, Identity, None)
} & 0.339 & 0.683 & 0.519 \\
  \cmidrule{2-6}
 & 3 & \tabincell{l}{Block 1: (1$\times$1 Conv, GeLU, None)\\Block 2: (1$\times3$ Conv, ReLU, LayerNorm)\\Block 3: (4-Head Self Attention, ReLU, LayerNorm)\\Block 4: (2-Head Self Attention, GeLU, None)\\Block 5: Zero \\Block 6: (2-Head Self Attention, Swish, None)
} & 0.337 & 0.681 & 0.517 \\
  \cmidrule{2-6}
  & 4 & \tabincell{l}{Block 1: Zero \\Block 2: (1$\times$7 Dconv, GeLU, LayerNorm)\\Block 3: Zero\\Block 4: (4-Head Self Attention, Swish, None)\\Block 5: (2-Head Self Attention, Swish, None)\\Block 6: (1$\times$5 Dconv, ReLU, None)
} & 0.335 & 0.678 & 0.515 \\
 \bottomrule
 \end{tabular}
}

\label{tab:seq_search}
\end{table}

\begin{table}[tp]
\centering
\caption{Searched interactions, MLPs and corresponding experimental results of interaction exploration (AMEIR-B) and additional aggregation MLP investigation (AMEIR-C) on Criteo and Avazu over 4 runs. 
Each interaction is represented by a tuple of non-sequential feature fields.
Each MLP includes a list of hidden layers, where each hidden layer is represented by a tuple of (hidden size, activation operation).}
\resizebox{\linewidth}{!}{
\begin{tabular}{m{1.3cm}<{\centering}|m{1.2cm}<{\centering}|m{7.5cm}<{\centering}|m{1.8cm}<{\centering}|m{4cm}<{\centering}|m{1.3cm}<{\centering}|m{1.3cm}<{\centering}}
 \toprule
Dataset & Run ID & Searched Interactions & Model & MLP & AUC & Log Loss \\
  \midrule
\multirow{16}{*}{Criteo} & \multirow{4}{*}{1} & \multirow{4}{7.5cm}{[(C10, C13), (C13, C26), (C15, C24), (C15, C20), (C11, C21), (C13, C19), (C4, C16), (C2, C15), (C7, C19), (C21, C26), (C19, C24), (C15, C19), (C14, C23), (C3, C6), (C10, C16)]} & \multirow{2}{*}{AMEIR-B} & \multirow{2}{4cm}{[(200, ReLU), (200, ReLU), (200, ReLU)]} & \multirow{2}{*}{0.8029} & \multirow{2}{*}{0.4487}\\
 & & & & & & \\
  \cmidrule{4-7}
 & & & \multirow{2}{*}{AMEIR-C} & \multirow{2}{4cm}{[(174, Swish), (21, ReLU), (21, Identity)]} & \multirow{2}{*}{0.8034} & \multirow{2}{*}{0.4482}\\
 & & & & & & \\
 \cmidrule{2-7}
 & \multirow{4}{*}{2} & \multirow{4}{7.5cm}{[(C6, C13), (C11, C24), (C4, C26), (C13, C26), (C2, C24), (C11, C20), (C6, C16), (C7, C16), (C10, C19), (C2, C15), (C3, C18), (C10, C24), (C4, C25), (C3, C21), (C16, C17)]} & \multirow{2}{*}{AMEIR-B} & \multirow{2}{4cm}{[(200, ReLU), (200, ReLU), (200, ReLU)]} & \multirow{2}{*}{0.8030} & \multirow{2}{*}{0.4486}\\
 & & & & & & \\
 \cmidrule{4-7}
 & & & \multirow{2}{*}{AMEIR-C} & \multirow{2}{4cm}{[(152, ReLU), (87, Identity), (21, ReLU)]} & \multirow{2}{*}{0.8035} & \multirow{2}{*}{0.4483}\\
 & & & & & & \\
  \cmidrule{2-7}
 & \multirow{4}{*}{3} & \multirow{4}{7.5cm}{[(C6, C13), (C10, C13), (C7, C11), (C11, C24), (C4, C26), (C13, C26), (C14, C18), (C2, C24), (C13, C16), (C15, C24), (C12, C16), (C12, C13), (C6, C11), (C2, C13), (C15, C20)]} & \multirow{2}{*}{AMEIR-B} & \multirow{2}{4cm}{[(200, ReLU), (200, ReLU), (200, ReLU)]} & \multirow{2}{*}{0.8032} & \multirow{2}{*}{0.4483}\\
 & & & & & & \\
 \cmidrule{4-7}
 & & & \multirow{2}{*}{AMEIR-C} & \multirow{2}{4cm}{[(152, ReLU), (65, ReLU), (21, ReLU)]} & \multirow{2}{*}{0.8036} & \multirow{2}{*}{0.4480}\\
 & & & & & & \\
  \cmidrule{2-7}
 & \multirow{4}{*}{4} & \multirow{4}{7.5cm}{[(C6, C13), (C13, C26), (C14, C18), (C12, C16), (C15, C20), (C11, C23), (C13, C19), (C3, C12), (C2, C15), (C10, C24), (C19, C24), (C2, C10), (C3, C15), (C4, C25), (C10, C16)]} & \multirow{2}{*}{AMEIR-B} & \multirow{2}{4cm}{[(200, ReLU), (200, ReLU), (200, ReLU)]} & \multirow{2}{*}{0.8028} & \multirow{2}{*}{0.4490}\\
 & & & & & & \\
 \cmidrule{4-7}
 & & & \multirow{2}{*}{AMEIR-C} & \multirow{2}{4cm}{[(174, Swish), (65, Dice), (21, Dice)]} & \multirow{2}{*}{0.8032} & \multirow{2}{*}{0.4485}\\
 & & & & & & \\
\midrule
\multirow{26}{*}{Avazu} & \multirow{6}{*}{1} & \multirow{6}{7.5cm}{[(C15, C18), (device\_conn\_type, C18), (app\_category, device\_type), (site\_category, C18), (app\_domain, C15), (site\_id, C21), (app\_category, C15), (site\_id, device\_id), (app\_id, C21), (C1, C15), (weekday, site\_id), (app\_domain, C17), (device\_id, C15), (day, app\_category), (day, device\_model)]} & \multirow{3}{*}{AMEIR-B} & \multirow{3}{4cm}{[(500, ReLU), (500, ReLU), (500, ReLU), (500, ReLU), (500, ReLU)]} & \multirow{3}{*}{0.7794} & \multirow{3}{*}{0.3796}\\
 & & & & & & \\
 & & & & & & \\
  \cmidrule{4-7}
 & & & \multirow{4}{*}{AMEIR-C} & \multirow{4}{4cm}{[(1092, Swish), (780, ReLU), (156, Identity), (156, ReLU), (156, Identity)]} & \multirow{4}{*}{0.7798} & \multirow{4}{*}{0.3794}\\
 & & & & & & \\
 & & & & & & \\
  & & & & & & \\
\cmidrule{2-7}
 & \multirow{6}{*}{2} & \multirow{6}{7.5cm}{[(device\_conn\_type, C18), (C1, C21), (site\_domain, device\_conn\_type), (day, C1), (site\_id, C21), (app\_category, C15), (site\_id, C20), (device\_model, C16), (site\_category, C21), (site\_category, device\_id), (weekday, site\_id), (C18, C19), (site\_id, C17), (app\_domain, C17), (device\_id, C15)]} & \multirow{3}{*}{AMEIR-B} & \multirow{3}{4cm}{[(500, ReLU), (500, ReLU), (500, ReLU), (500, ReLU), (500, ReLU)]} & \multirow{3}{*}{0.7796} & \multirow{3}{*}{0.3795}\\
 & & & & & & \\
 & & & & & & \\
  \cmidrule{4-7}
 & & & \multirow{3}{*}{AMEIR-C} & \multirow{3}{4cm}{[(936, Identity), (468, Dice), (156, ReLU), (156, Swish), (156, ReLU)]} & \multirow{3}{*}{0.7799} & \multirow{3}{*}{0.3793}\\
 & & & & & & \\
 & & & & & & \\
\cmidrule{2-7}
 & \multirow{6}{*}{3} & \multirow{6}{7.5cm}{[(C15, C18), (day, C16), (banner\_pos, app\_id), (C1, C21), (app\_id, device\_type), (device\_id, C16), (banner\_pos, device\_type), (site\_category, device\_type), (banner\_pos, C20), (app\_id, device\_id), (site\_id, device\_id), (app\_id, C21), (C1, C15), (weekday, site\_id), (device\_id, C15)]} & \multirow{3}{*}{AMEIR-B} & \multirow{3}{4cm}{[(500, ReLU), (500, ReLU), (500, ReLU), (500, ReLU), (500, ReLU)]} & \multirow{3}{*}{0.7802} & \multirow{3}{*}{0.3793}\\
 & & & & & & \\
 & & & & & & \\
  \cmidrule{4-7}
 & & & \multirow{3}{*}{AMEIR-C} & \multirow{3}{4cm}{[(936, ReLU), (468, Identity), (156, Identity), (156, ReLU), (156, ReLU)]
} & \multirow{3}{*}{0.7806} & \multirow{3}{*}{0.3791}\\
 & & & & & & \\
 & & & & & & \\
\cmidrule{2-7}
 & \multirow{8}{*}{4} & \multirow{8}{7.5cm}{[(device\_conn\_type, C18), (app\_category, device\_type), (day, C1), (device\_id, C16), (weekday, site\_domain), (banner\_pos, C20), (app\_domain, C15), (day, weekday), (site\_id, C20), (site\_category, device\_conn\_type), (hour, C14), (app\_domain, device\_ip), (site\_domain, device\_ip), (weekday, site\_id), (C16, C21)]} & \multirow{4}{*}{AMEIR-B} & \multirow{4}{4cm}{[(500, ReLU), (500, ReLU), (500, ReLU), (500, ReLU), (500, ReLU)]} & \multirow{4}{*}{0.7797} & \multirow{4}{*}{0.3792}\\
 & & & & & & \\
 & & & & & & \\
 & & & & & & \\
  \cmidrule{4-7}
 & & & \multirow{4}{*}{AMEIR-C} & \multirow{4}{4cm}{[(936, ReLU), (312, ReLU), (156, Swish), (156, ReLU), (156, Identity)]
} & \multirow{4}{*}{0.7802} & \multirow{4}{*}{0.3790}\\
 & & & & & & \\
 & & & & & & \\
  & & & & & & \\
 \bottomrule
 \end{tabular}
}

\label{tab:nonseq}
\end{table}

\end{document}